\documentclass[conference]{IEEEtran}

\usepackage{amsmath,amsfonts,bm}
\usepackage{xparse}

\DeclareDocumentCommand\W{ g g }{%
        \IfNoValueTF {#1} {\mathbf{W}} {
            \IfNoValueTF {#2} {\mathbf{W}^{(#1)}}{\mathbf{W}^{(#1)}_{#2}}
        }
}

\DeclareDocumentCommand\bias{ g g }{%
        \IfNoValueTF {#1} {\mathbf{b}} {
            \IfNoValueTF {#2} {\mathbf{b}^{(#1)}}{\mathbf{b}^{(#1)}_{#2}}
        }
}

\DeclareDocumentCommand\betavar{ g g }{%
        \IfNoValueTF {#1} {\bm{\beta}} {
            \IfNoValueTF {#2} {{\bm{\beta}^{(#1)}}{}}{\bm{\beta}^{(#1)}_{#2}}
        }
}

\DeclareDocumentCommand\xivar{ g g }{%
        \IfNoValueTF {#1} {\bm{\xi}} {
            \IfNoValueTF {#2} {{\bm{\xi}^{(#1)}}{}}{\bm{\xi}^{(#1)}_{#2}}
        }
}

\DeclareDocumentCommand\xivarn{ g g }{%
        \IfNoValueTF {#1} {\bm{\xi^-}} {
            \IfNoValueTF {#2} {\bm{\xi^-}^{+(#1)}}{\bm{\xi^-}^{+(#1)}_{#2}}
        }
}

\DeclareDocumentCommand\xivarp{ g g }{%
        \IfNoValueTF {#1} {\bm{\xi^+}} {
            \IfNoValueTF {#2} {\bm{\xi^+}^{+(#1)}}{\bm{\xi^+}^{+(#1)}_{#2}}
        }
}

\DeclareDocumentCommand\nuvar{ g g }{%
        \IfNoValueTF {#1} {{\bm{\nu}}} {
            \IfNoValueTF {#2} {{\bm{\nu}^{(#1)}}{}}{\nu^{(#1)}_{#2}{}}
        }
}

\DeclareDocumentCommand\hnuvar{ g g }{%
        \IfNoValueTF {#1} {\bm{\hat{\nu}}} {
            \IfNoValueTF {#2} {{\bm{\hat{\nu}}^{(#1)}}{}}{\hat{\nu}^{(#1)}_{#2}{}}
        }
}

\DeclareDocumentCommand\muvar{ g g }{%
        \IfNoValueTF {#1} {\bm{\mu}} {
            \IfNoValueTF {#2} {{\bm{\mu}^{(#1)}}{}}{\mu^{(#1)}_{#2}}
        }
}

\DeclareDocumentCommand\tauvar{ g g }{%
        \IfNoValueTF {#1} {\bm{\tau}} {
            \IfNoValueTF {#2} {{\bm{\tau}^{(#1)}}{}}{\tau^{(#1)}_{#2}}
        }
}

\DeclareDocumentCommand\pivar{ g g }{%
        \IfNoValueTF {#1} {\bm{\pi}} {
            \IfNoValueTF {#2} {{\bm{\pi}^{(#1)}}{}}{\pi^{(#1)}_{#2}}
        }
}

\DeclareDocumentCommand\gammavar{ g g }{%
        \IfNoValueTF {#1} {\bm{\gamma}} {
            \IfNoValueTF {#2} {{\bm{\gamma}^{(#1)}}{}}{\gamma^{(#1)}_{#2}}
        }
}

\DeclareDocumentCommand\lambdavar{ g g }{%
        \IfNoValueTF {#1} {\bm{\lambda}} {
            \IfNoValueTF {#2} {{\bm{\lambda}^{(#1)}}{}}{\lambda^{(#1)}_{#2}}
        }
}

\DeclareDocumentCommand\tbetavar{ g g }{%
        \IfNoValueTF {#1} {{\bm{\tilde{\beta}}}} {
            \IfNoValueTF {#2} {{{\bm{\tilde{\beta}}}^{(#1)}}{}}{{{\tilde{\beta}}^{(#1)}_{#2}}}
        }
}

\DeclareDocumentCommand\alphavar{ g g }{%
        \IfNoValueTF {#1} {\bm{\alpha}} {
            \IfNoValueTF {#2} {{\bm{\alpha}^{(#1)}}}{\alpha^{(#1)}_{#2}}
        }
}

\DeclareDocumentCommand\hfunc{ g g }{%
        \IfNoValueTF {#1} {h} {
            \IfNoValueTF {#2} {{{h}^{(#1)}}}{h^{(#1)}_{#2}}
        }
}

\DeclareDocumentCommand\zcut{ g g }{%
        \IfNoValueTF {#1} {\bm{q}} {
            \IfNoValueTF {#2} {{\bm{q}^{(#1)}}}{q^{(#1)}_{#2}}
        }
}

\DeclareDocumentCommand\Zcut{ g g }{%
        \IfNoValueTF {#1} {\bm{Q}} {
            \IfNoValueTF {#2} {{\bm{Q}^{(#1)}}}{\bm{Q}^{(#1)}_{#2}}
        }
}

\DeclareDocumentCommand\xcut{ g g }{%
        \IfNoValueTF {#1} {\bm{h}} {
            \IfNoValueTF {#2} {{\bm{h}^{(#1)}}}{h^{(#1)}_{#2}}
        }
}

\DeclareDocumentCommand\Xcut{ g g }{%
        \IfNoValueTF {#1} {\bm{H}} {
            \IfNoValueTF {#2} {{\bm{H}^{(#1)}}}{\bm{H}^{(#1)}_{#2}}
        }
}

\DeclareDocumentCommand\hxcut{ g g }{%
        \IfNoValueTF {#1} {\bm{g}} {
            \IfNoValueTF {#2} {{\bm{g}^{(#1)}}}{g^{(#1)}_{#2}}
        }
}

\DeclareDocumentCommand\hXcut{ g g }{%
        \IfNoValueTF {#1} {\bm{G}} {
            \IfNoValueTF {#2} {{\bm{G}^{(#1)}}}{\bm{G}^{(#1)}_{#2}}
        }
}

\DeclareDocumentCommand\D{ g g }{%
        \IfNoValueTF {#1} {\mathbf{D}} {
            \IfNoValueTF {#2} {\mathbf{D}^{(#1)}}{\mathbf{D}^{(#1)}_{#2}}
        }
}

\DeclareDocumentCommand\A{ g g }{%
        \IfNoValueTF {#1} {\mathbf{A}} {
            \IfNoValueTF {#2} {\mathbf{A}^{(#1)}}{\mathbf{A}^{(#1)}_{#2}}
        }
}

\DeclareDocumentCommand\a{ g g }{%
        \IfNoValueTF {#1} {\mathbf{a}} {
            \IfNoValueTF {#2} {\mathbf{a}^{(#1)}}{{a}^{(#1)}_{#2}}
        }
}

\DeclareDocumentCommand\al{ g g }{%
        \IfNoValueTF {#1} {\underline{\mathbf{a}}} {
            \IfNoValueTF {#2} {\underline{\mathbf{a}}^{(#1)}}{\underline{a}^{(#1)}_{#2}}
        }
}

\DeclareDocumentCommand\au{ g g }{%
        \IfNoValueTF {#1} {\overline{\mathbf{a}}} {
            \IfNoValueTF {#2} {\overline{\mathbf{a}}^{(#1)}}{\overline{a}^{(#1)}_{#2}}
        }
}

\DeclareDocumentCommand\c{ g }{%
        \IfNoValueTF {#1} {\bm{c}} {
            {c^{({#1})}}
        }
}

\DeclareDocumentCommand\cl{ g }{%
        \IfNoValueTF {#1} {\underline{c}} {
            {\underline{c}^{({#1})}}
        }
}

\DeclareDocumentCommand\cu{ g }{%
        \IfNoValueTF {#1} {\overline{c}} {
            {\overline{c}^{({#1})}}
        }
}

\DeclareDocumentCommand\AA{ g g }{
        \IfNoValueTF {#1} {\mathbf{\Omega}} {
            \IfNoValueTF {#2} {\mathbf{\Omega}(#1, #1)}{\mathbf{\Omega}(#1, #2)}
        }
}

\DeclareDocumentCommand\S{ g g }{%
        \IfNoValueTF {#1} {\mathbf{S}} {
            \IfNoValueTF {#2} {\mathbf{S}^{(#1)}}{\mathbf{S}^{(#1)}_{#2}}
        }
}

\DeclareDocumentCommand\K{ g g }{%
        \IfNoValueTF {#1} {\mathbf{K}} {
            \IfNoValueTF {#2} {\mathbf{K}^{(#1)}}{\mathbf{K}^{(#1)}_{#2}}
        }
}

\DeclareDocumentCommand\B{ g g }{%
        \IfNoValueTF {#1} {\mathbf{B}} {
            \IfNoValueTF {#2} {\mathbf{B}^{(#1)}}{\mathbf{B}^{(#1)}_{#2}}
        }
}

\DeclareDocumentCommand\lowerb{ g g }{%
        \IfNoValueTF {#1} {{\mathbf{\underline{b}}}} {
            \IfNoValueTF {#2} {{\mathbf{\underline{b}}}^{(#1)}}{{\mathbf{\underline{b}}}^{(#1)}_{#2}}
        }
}

\DeclareDocumentCommand\z{ g g }{%
        \IfNoValueTF {#1} {\mathbf{z}} {
            \IfNoValueTF {#2} {\mathbf{z}^{(#1)}}{z^{(#1)}_{#2}}
        }
}

\DeclareDocumentCommand\hz{ g g }{%
        \IfNoValueTF {#1} {\hat{\mathbf{z}}} {
            \IfNoValueTF {#2} {\hat{\mathbf{z}}^{(#1)}}{\hat{z}^{(#1)}_{#2}}
        }
}

\DeclareDocumentCommand\hx{ g g }{%
        \IfNoValueTF {#1} {\hat{\boldsymbol{x}}} {
            \IfNoValueTF {#2} {\hat{\boldsymbol{x}}^{(#1)}}{\hat{x}^{(#1)}_{#2}}
        }
}

\DeclareDocumentCommand\x{ g g }{%
        \IfNoValueTF {#1} {\boldsymbol{x}} {
            \IfNoValueTF {#2} {\boldsymbol{x}^{(#1)}}{x^{(#1)}_{#2}}
        }
}

\DeclareDocumentCommand\bu{ g g }{%
        \IfNoValueTF {#1} {\boldsymbol{u}} {
            \IfNoValueTF {#2} {\boldsymbol{u}^{(#1)}}{{u}^{(#1)}_{#2}}
        }
}

\DeclareDocumentCommand\buh{ g g }{%
        \IfNoValueTF {#1} {\boldsymbol{u}^*} {
            \IfNoValueTF {#2} {\boldsymbol{{u}}^{*(#1)}}{{{u}}^{*(#1)}_{#2}}
        }
}

\DeclareDocumentCommand\bl{ g g }{%
        \IfNoValueTF {#1} {\boldsymbol{l}} {
            \IfNoValueTF {#2} {\boldsymbol{l}^{(#1)}}{{l}^{(#1)}_{#2}}
        }
}

\DeclareDocumentCommand\blh{ g g }{%
        \IfNoValueTF {#1} {\boldsymbol{l}^*} {
            \IfNoValueTF {#2} {\boldsymbol{l}^{*(#1)}}{{{l}}^{*(#1)}_{#2}}
        }
}

\DeclareDocumentCommand\aaa{ g }{%
        \IfNoValueTF {#1} {\bm{a}} {
            {\bm{a}^{({#1})}}
        }
}

\DeclareDocumentCommand\haaa{ g }{%
        \IfNoValueTF {#1} {\bm{\hat{a}}} {
            {\bm{\hat{a}}^{({#1})}}
        }
}

\DeclareDocumentCommand\bbb{ g g }{%
        \IfNoValueTF {#1} {\mathbf{P}} {
            \IfNoValueTF {#2} {{\mathbf{P}_{#1}}}{{\mathbf{P}_{#1}^{({#2})}}}
        }
}

\DeclareDocumentCommand\hbbb{ g g }{%
        \IfNoValueTF {#1} {\mathbf{\hat{P}}} {
            \IfNoValueTF {#2} {{\mathbf{\hat{P}}_{#1}}}{{\mathbf{\hat{P}}_{#1}^{({#2})}}}
        }
}

\DeclareDocumentCommand\ccc{ g g }{%
        \IfNoValueTF {#1} {\mathbf{q}} {
            \IfNoValueTF {#2} {{\mathbf{q}_{#1}}}{{\mathbf{q}_{#1}^{(#2)}}{}}
        }
}

\DeclareDocumentCommand\constc{ g }{%
        \IfNoValueTF {#1} {c} {
            {c^{({#1})}}
        }
}

\DeclareDocumentCommand\setz{ g g }{%
        \IfNoValueTF {#1} {\mathcal{Z}} {
            \IfNoValueTF {#2} {\mathcal{Z}^{(#1)}}{\mathcal{Z}^{(#1)}_{#2}}
        }
}

\DeclareDocumentCommand\setzp{ g g }{%
        \IfNoValueTF {#1} {\mathcal{Z^+}} {
            \IfNoValueTF {#2} {\mathcal{Z}^{+(#1)}}{\mathcal{Z}^{+(#1)}_{#2}}
        }
}

\DeclareDocumentCommand\setzn{ g g }{%
        \IfNoValueTF {#1} {\mathcal{Z^-}} {
            \IfNoValueTF {#2} {\mathcal{Z}^{-(#1)}}{\mathcal{Z}^{-(#1)}_{#2}}
        }
}

\DeclareDocumentCommand\tsetz{ g g }{%
        \IfNoValueTF {#1} {\tilde{\mathcal{Z}}} {
            \IfNoValueTF {#2} {\tilde{\mathcal{Z}}^{(#1)}}{\tilde{\mathcal{Z}}^{(#1)}_{#2}}
        }
}

\DeclareDocumentCommand\seti{ g g }{%
        \IfNoValueTF {#1} {\mathcal{I}} {
            \IfNoValueTF {#2} {\mathcal{I}^{(#1)}}{\mathcal{I}^{(#1)}_{#2}}
        }
}

\DeclareDocumentCommand\setip{ g g }{%
        \IfNoValueTF {#1} {\mathcal{I}^{+}} {
            \IfNoValueTF {#2} {\mathcal{I}^{+(#1)}}{\mathcal{I}^{+(#1)}_{#2}}
        }
}

\DeclareDocumentCommand\setin{ g g }{%
        \IfNoValueTF {#1} {\mathcal{I}^{-}} {
            \IfNoValueTF {#2} {\mathcal{I}^{-(#1)}}{\mathcal{I}^{-(#1)}_{#2}}
        }
}

\DeclareDocumentCommand\tseti{ g g }{%
        \IfNoValueTF {#1} {\tilde{\mathcal{I}}} {
            \IfNoValueTF {#2} {\tilde{\mathcal{I}}^{(#1)}}{\tilde{\mathcal{I}}^{(#1)}_{#2}}
        }
}

\DeclareDocumentCommand\tz{ g g }{%
        \IfNoValueTF {#1} {\tilde{z}} {
            \IfNoValueTF {#2} {\tilde{z}^{(#1)}}{\tilde{z}^{(#1)}_{#2}}
        }
}

\DeclareDocumentCommand\f{ g g }{%
        \IfNoValueTF {#1} {f} {
            \IfNoValueTF {#2} {f^{(#1)}}{f^{(#1)}_{#2}}
        }
}

\DeclareDocumentCommand\lf{ g g }{%
        \IfNoValueTF {#1} {\underline{f}} {
            \IfNoValueTF {#2} {\underline{f}^{(#1)}}{\underline{f}^{(#1)}_{#2}}
        }
}

\def\eqref#1{(\ref{#1})}

\def\1{\bm{1}}

\def\vb{{\bm{b}}}

\def\vp{{\bm{p}}}

\DeclareMathAlphabet{\mathsfit}{\encodingdefault}{\sfdefault}{m}{sl}
\SetMathAlphabet{\mathsfit}{bold}{\encodingdefault}{\sfdefault}{bx}{n}

\def\gL{{\mathcal{L}}}

\def\sM{{\mathbb{M}}}

\def\sP{{\mathbb{P}}}

\def\sS{{\mathbb{S}}}

\def\sX{{\mathbb{X}}}
\def\sY{{\mathbb{Y}}}

\usepackage{times}

\usepackage[numbers]{natbib}
\usepackage{multicol}
\usepackage[bookmarks=true]{hyperref}
\usepackage{fancyhdr}

\usepackage{amsmath}
\usepackage{algorithm}
\usepackage{algorithmic}
\usepackage{graphicx}
\usepackage{array}
\usepackage{stfloats}
\usepackage{subcaption}
\usepackage{booktabs}
\usepackage{adjustbox}

\usepackage{color}
\usepackage{soul}
\newcommand{\rev}[1]{{\color{black}{#1}}}

\fancypagestyle{firstpagestyle}{
    \fancyhf{}
    \fancyhead[C]{\Large\href{https://roboticsconference.org/2024/program/papers/}{Accepted to Robotics: Science and Systems (RSS) 2024}}
}

\begin{document}

\title{Dynamic Adversarial Attacks on \\ Autonomous Driving Systems}

\author{Amirhosein Chahe ~ Chenan Wang ~ Abhishek Jeyapratap ~ Kaidi Xu ~ Lifeng Zhou\\
Drexel University\\
{\tt\small \{ac4462, cw3344, aj928, kx46, lz457\}@drexel.edu}
}


\maketitle

\thispagestyle{firstpagestyle}

\begin{abstract}
This paper introduces an attacking mechanism to challenge the resilience of autonomous driving systems. Specifically, we manipulate the decision-making processes of an autonomous vehicle by dynamically displaying adversarial patches on a screen mounted on another moving vehicle. These patches are optimized to deceive the object detection models into misclassifying targeted objects, e.g., traffic signs. Such manipulation has significant implications for critical multi-vehicle interactions such as intersection crossing, which are vital for safe and efficient autonomous driving systems. 
Particularly, we make four major contributions. First, we introduce a novel adversarial attack approach where the patch is not co-located with its target, enabling more versatile and stealthy attacks. Moreover, our method utilizes dynamic patches displayed on a screen, allowing for adaptive changes and movements, enhancing the flexibility and performance of the attack. To do so, we design a Screen Image Transformation Network (SIT-Net), which simulates environmental effects on the displayed images, narrowing the gap between simulated and real-world scenarios. Further, we integrate a positional loss term into the adversarial training process to increase the success rate of the dynamic attack. Finally, we shift the focus from merely attacking perceptual systems to influencing the decision-making algorithms of self-driving systems. Our experiments demonstrate the first successful implementation of such dynamic adversarial attacks in real-world autonomous driving scenarios, paving the way for advancements in the field of robust and secure autonomous driving. 
\end{abstract}

\IEEEpeerreviewmaketitle

\section{Introduction}
\label{sec:intro}
As autonomous vehicles and driver assistance systems become more prevalent, it is imperative to ensure that their decision-making systems are robust. The ability to detect and recognize traffic signs and cars accurately is crucial for autonomous vehicles to make safe and efficient decisions~\cite{VANBRUMMELEN2018384,Geng2020,wang2023does}. 
Deep neural networks (DNNs)~\cite{DNN} provide an efficient framework for real-time object recognition widely used by autonomous driving systems. However, it is not immune to adversarial attacks, which can potentially have serious consequences in real-world scenarios~\cite{Szegedy2013, Nguyen2018, xu2018structured,zhao2019admm}. In the past several years, scientists have investigated the reasons behind neural networks' vulnerability to adversarial examples and devised techniques for generating digitally perturbed images~\cite{FawziMF16, LuIF17, Sabour2015, Moosavi-Dezfooli16, Chen21}. In a range of real-world scenarios, 
several studies indicated that adversarial perturbations on actual objects can deceive DNN-based object detectors in real-world scenarios~\cite{Wang2021,Huang2019,Du_2022_WACV,Brown2017AdversarialP,AthalyeS17,Xu2020,Sharif2016,Cao2019,Juncheng2019}. For instance, To avoid detection, an individual can either carry a cardboard plate with a printed adversarial patch as described in~\cite{Thys2019} or wear an adversarial T-shirt as proposed by~\cite{Xu2020}, both oriented towards the surveillance camera. Furthermore,~\cite{Chen2018} and~\cite{Eykholt18} have demonstrated the feasibility of real-world adversarial attacks on Yolo~\cite{redmon16} and Faster R-CNN~\cite{fRcnn}. They focus on manipulating stop signs by hiding the stop sign using an adversarial stop sign poster or by affixing adversarial stickers onto the sign. 

\begin{figure*}[htb]
    \centering
    \begin{subfigure}{\textwidth}
      \centering
      \setlength{\lineskip}{2pt}
    \includegraphics[width=0.24\linewidth]{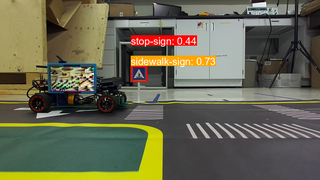}
    \includegraphics[width=0.24\linewidth]{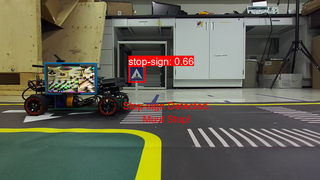}\hspace{0.0001\linewidth}
    \includegraphics[width=0.24\linewidth]{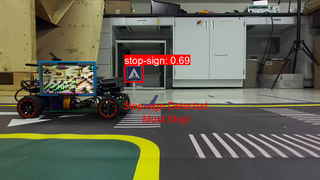}\hspace{0.0001\linewidth}
    \includegraphics[width=0.24\linewidth]{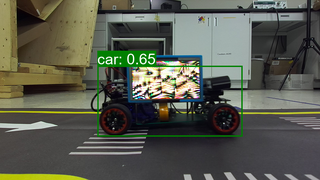}\\
    \includegraphics[width=0.24\linewidth]{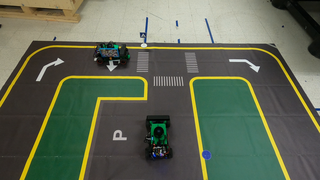}
    \includegraphics[width=0.24\linewidth]{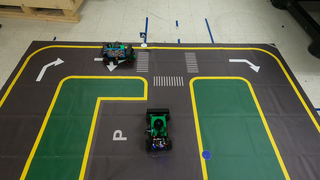}\hspace{0.0001\linewidth}
    \includegraphics[width=0.24\linewidth]{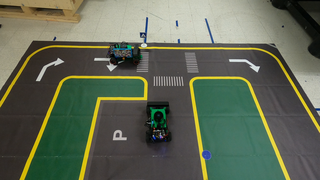}\hspace{0.0001\linewidth}
    \includegraphics[width=0.24\linewidth]{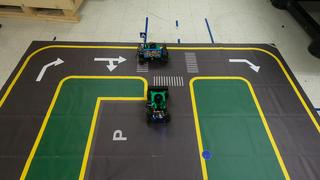}
    \caption{Adversarial patch attacks on the Pedestrian sign.}
    \end{subfigure}
    
    \vspace{0.5em}
    
    \begin{subfigure}{\textwidth}
      \centering
      \setlength{\lineskip}{2pt}
      \includegraphics[width=0.24\linewidth]{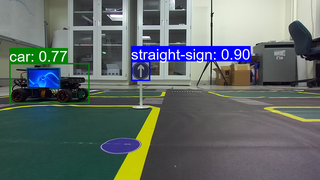}\hspace{0.0001\linewidth}
      \includegraphics[width=0.24\linewidth]{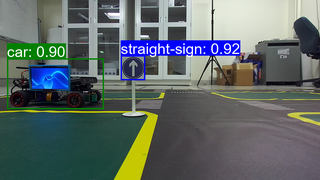}\hspace{0.0001\linewidth}
      \includegraphics[width=0.24\linewidth]{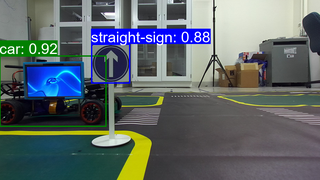}\hspace{0.0001\linewidth}
      \includegraphics[width=0.24\linewidth]{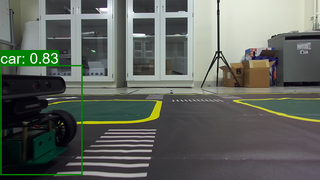}\\
      \includegraphics[width=0.24\linewidth]{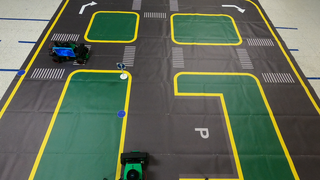}\hspace{0.0001\linewidth}
      \includegraphics[width=0.24\linewidth]{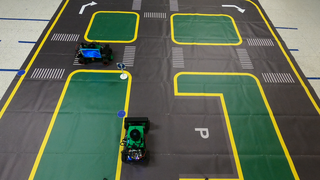}\hspace{0.0001\linewidth}
      \includegraphics[width=0.24\linewidth]{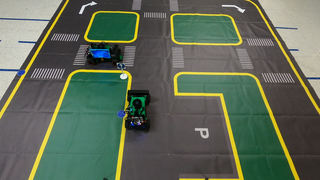}\hspace{0.0001\linewidth}
      \includegraphics[width=0.24\linewidth]{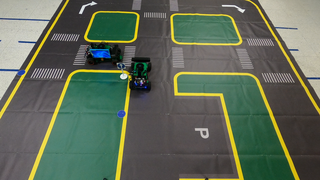}
      \caption{Benign trial.}
    \end{subfigure}
    \caption{ Sequential frames of a multi-robot interaction in an intersection: The top row of (a) demonstrates a dynamic adversarial patch attack, where the patch is displayed on the screen of the \texttt{patch car}. The patch is designed to mislead the \texttt{camera car}'s perception system, i.e., the camera car thinks the Pedestrian sign is a Stop sign. The top row of (b) illustrates the system's correct behavior during a benign trial without adversarial interference. The bottom rows of (a) and (b) are the top views of the settings.
    }
    \label{fig:realworld test}
\end{figure*}

However, in~\cite{Jiajun17}, it showed that the printed patches cannot fool object detectors across a wide range of distances and angles. To come up with a variety of distances and angles, we aim to deceive and mislead the DNN detectors by generating dynamic adversarial patches which are adapted based on the camera and target positions. We place the patches in a location distinct from the location of the target object. Specifically, we display the patch using a screen mounted on a robot car rather than on the target traffic sign itself. Our methodology employs an optimization process that operates at the pixel level, aiming to display a patch that significantly enhances the probability of misclassifying the object within a substantial dataset. Notably, our attack is carried out by a dynamic patch displayed on the screen, making its formulation and simulation more challenging due to several environmental variables. 

The goal of this work is to affect the decision-making of an autonomous vehicle by affixing a custom-designed dynamic patch to another moving vehicle, causing the object detector model to misclassify an object, e.g., a non-restrictive traffic sign as a restrictive one as shown in Figure~\ref{fig:realworld test}.  This adversarial attack could potentially manipulate the behavior of the attacked vehicle in various vehicle interaction scenarios including navigating intersections, changing lanes, and merging into a lane, underpinning the importance of robust object detection in safety-critical autonomous driving systems.  

\noindent \textbf{Contributions.} In this paper, we make four main contributions as follows. 
\begin{itemize}
    \item We formulate the problem of attacking the object detection of traffic signs by displaying a dynamic patch on a screen mounted on an autonomous robot car which enables the patch not only to be dynamically changing and moving but also short-lived and harder to detect. 
    \item We reformulate the loss function of state-of-the-art adversarial neural network models by integrating a positional loss term that aligns the target object bounding box to increase the likelihood of a successful attack in the presence of corroborating sensory data.
    \item We propose a SIT-Net that simulates the color and contrast transformation of the displayed image on the screen captured by the camera to mitigate the gaps between the simulation and the physical world. 
    \item We conduct extensive experiments to show successful attacks across various multi-vehicle interaction scenarios passing the intersection, and attacking various traffic signs including Go-straight, Turn, and Pedestrian signs. This allows the attacker to adversarially manipulate other vehicles' decisions to take the lead and move with lower risk.  
\end{itemize}
Furthermore, to the best of our knowledge, \textit{we are the first to design and demonstrate dynamic adversarial attacks on real-world autonomous driving systems}.


\section{Related Work}
In the field of autonomous vehicles, the authors in~\cite{Zhao2019} proposed systematic solutions for Hiding Attack (HA) and Appearing Attack (AA), enhancing the robustness of adversarial examples (AEs) against varying distances and angles. In~\cite{Jia2022}, a method for generating physical AEs that fools traffic sign recognition of autonomous vehicles was presented which employs single and multiple bounding box filters, as well as considering different attack vectors, such as hiding, appearance, non-target, and target attacks. In~\cite{Lovisotto2020SLAPIP}, the authors proposed Short-Lived Adversarial Perturbations (SLAP) that allow adversaries to create physically robust real-world AEs using a projector that provides better control over the attack compared to traditional adversarial patches, as projections can be turned on and off as needed and leave no obvious trace of an attack. Furthermore, in~\cite{DBLP-shahar}, a dynamic adversarial patch technique was introduced to circumvent the object detection model by employing several dynamic patches applied to a target object that adapts to the location of the camera. The authors showed by switching between optimized patches based on the camera's position, the attack can adapt to different situations and achieve optimal results. 

In all the previous physical targeted attacks, the patch is deployed at the location of the target object, which makes the patch implementation hard and or impossible in most practical cases. In this paper, we focus on the AEs in which a patch is in a different place from the target object and is dynamically displayed through the screen mounted on a either moving or stationary vehicle. Moreover, the final goal is to mislead the autonomous driving system's decision-making instead of purely deceiving its perceptions.

\section{Problem Formulation}

We investigate physical dynamic adversarial attacks on a DNN object detector for various autonomous driving scenarios where two robot cars, namely \texttt{patch car} and the \texttt{camera car},  are interacting with each other. The \texttt{camera car} employs a DNN object detector to detect other cars and traffic signs, which informs its decision-making process. Additionally, it utilizes a range of sensors, such as Lidar, to acquire the positions of these objects. The \texttt{patch car} is equipped with a screen that displays a deceptive patch. We aim to trick the \texttt{camera car}'s decision making such as coming to a halt by attacking its object detector, thereby allowing the \texttt{patch car} to take over and navigate through with earlier passing time and lower risk. Importantly, we also consider the camera car's ability to cross-reference detected object positions with its proximity data. For example, the agent matches detected objects' locations with the Lidar point clouds to verify it for the decision-making process~\cite{Xie2020, ASVADI201820}. This adds an extra layer of complexity to the task of successfully deceiving it. In the following, we introduce our attack model in two dimensions---the attacker and the objective.
\subsection{Attacker}
Attacker, i.e., the \texttt{patch car}, is a specialized agent that can display adversarial patches dynamically on a display screen attached to it. Instead of altering objects physically, this agent aims to manipulate object recognition processes through visual perturbations shown on its screen. The attacker employs a white-box attack strategy, granting it full access to the object detector model. It can also obtain the proximity of the objects including the position of the \texttt{camera car} using a Lidar.

To formally define the dynamic patch, let \( \sS \) be the state space representing all possible positions of the \texttt{camera car}, \( s_c \) and the \texttt{patch car}, \( s_p \), and \( \sP' \) be the space of all possible adversarial patches. The function \( T \) maps the patch \(\delta\) given a tuple of states \( (s_c, s_p) \) to a specific transformed adversarial patch \(\delta'\) in \( \sP' \). Moreover, to model environmental parameters affecting the patch visibility, we integrate the variable \(\phi \in \Phi\) into the map function.   This can be denoted as:
\begin{equation}
    T(\delta,s_{c},s_{p}, \phi): \sP \times \sS \times \sS \times \Phi \rightarrow \sP'
    \label{eq: delta}
\end{equation}
The variable \(\phi\) represents a specific color transformation of the displayed patch in the camera's image over all possible ones. We discuss it in the Section~\ref{sec: sitnet}.

\subsection{Objectives}
We aim to devise an attack that misguides the object detector model, causing it to incorrectly classify a traffic sign to a target sign when viewed from the perspective of the \texttt{camera car}.
We formally define the problem given that:
\begin{itemize}
    \item $\sX$: the space of all possible input images.
    \item $\sY$: the space of potential output predictions.
    \item $\sS$: the space of possible positions of the cars.
    \item $f: \sX \times \Theta \rightarrow \sY$: the detector model, parameterized by weights $\theta \in \Theta$.
\end{itemize}
The objective is to find the patch that maximizes the probability of the target class, e.g., stop-sign.
\begin{equation}
\begin{aligned}
& \underset{\delta}{\text{maximize}}
& & P[f(\pi(x,\delta,s_c,s_p), \theta) = (\vp_{\text{target}},\vb_{\text{target}}) ],\\
& \text{subject to}
& & \pi(x,\delta,s_c,s_p) = x + T(\delta,s_c,s_p,\phi)\\
& & &\; \forall x \in \sX, \; \forall s_c,s_p \in \sS, \; \forall \phi \in \Phi.
\end{aligned}
\end{equation}
$\pi(x,\delta,s_c,s_p)$ denotes the input image after applying the transformed patch. The patch must be applied within the coordination of the display screen which depends on the positions of the cars.
$T(\delta,s_{c},s_{p}, \phi)$ denotes the adversarial patch displayed on the screen transformation function provided earlier in Equation~\ref{eq: delta}.
$\vp_{\text{target}}$ denotes the output prediction corresponding to the confidence of the target. 
$\vb_{\text{target}}$ denotes the output prediction corresponding to the bounding box of the target.

 We utilize gradient descent optimization to identify the patch that maximizes the misclassification rate of the detector model. Gradient descent is widely used for training adversarial attacks on DNNs~\cite{Lee2019}. In our case, we aim to optimize the patch that minimizes the loss function, as described in the section~\ref{sec: obj-func}.

According to the proposed function~\ref{eq: delta} for the patch, it depends on the position of the cars ($s_p, s_c$) and the environmental parameters. In Section~\ref{sec: dyn-env}, we introduce how to capture the changes in the positions of the cars and integrate these, resulting in a dynamic patch. 
To integrate the environmental parameters $\Phi$, we propose a solution in Section~\ref{sec: sitnet}.

\subsection{Objective Function}\label{sec: obj-func}

To effectively deceive the \texttt{camera car}, our approach must address both the positional accuracy and the classification accuracy of the adversarial patch. The objective function $\mathcal{O}(\delta)$ is defined as follows:
\begin{equation}
    \mathcal{O}(\delta) = \mathbb{E}\left[ \vp_{\text{target}}(i) \cdot \vp_{\text{obj}}(i) \cdot \text{IoU}(\vb_{\text{target}}(i), \vb_{\text{orig}})\right], i \in \sM
\end{equation}
\begin{equation*}
     \delta^*=\min_\delta - \mathcal{O}(\delta).
     \label{eq: objective}
\end{equation*}
$\vp_{\text{target}}(i)$ is the probability of the $i$-th bounding box being the target class (e.g., stop sign). $\vp_{\text{obj}}(i)$ is the objectiveness probability of the $i$-th bounding box \rev{which is a likelihood that measures the confidence that the bounding box contains an object}. The
$\text{IoU}(\vb_{\text{target}}(i), \vb_{\text{orig}})$ is the Intersection over Union (IoU) between the $i$-th bounding box of the target class (e.g., stop sign). The  $\vb_{\text{target}}(i)$ is predicted by model and $\vb_{\text{orig}}$ is the original object bounding box. $\sM$ is the set of all the bounding boxes overlapping the original object. $\delta^*$ is the optimized patch. The objective function, \(\mathcal{O}\), is designed to optimize two key factors: 

\noindent\textbf{Confidence score.} The function aims to increase the probability that the object detector misclassifies the objects to the target class (e.g., a stop sign). This involves manipulating the patch in such a way that the object detector is highly confident in its misclassification.

\noindent\textbf{Bounding box overlap.} Alongside the confidence score, the function also aims to maximize the overlap between the bounding box predicted by the object detector caused by the adversarial patch and the bounding box of the actual object. 
The combination of these two factors ensures that the adversarial patch not only deceives the detector in terms of classification but also aligns accurately with the expected position of the target object, as determined by the \texttt{camera car}'s other sensory inputs, e.g., Lidar.

\begin{figure}[htp]
\setlength{\lineskip}{0pt}
\centering
\includegraphics[width=0.19\linewidth]{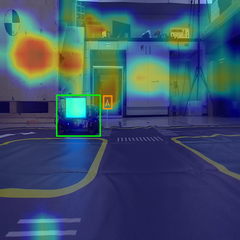}\hspace{0.0001\linewidth}
\includegraphics[width=0.19\linewidth]{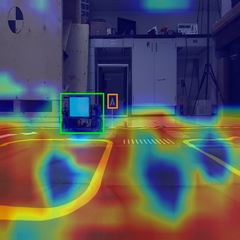}\hspace{0.0001\linewidth}
\includegraphics[width=0.19\linewidth]{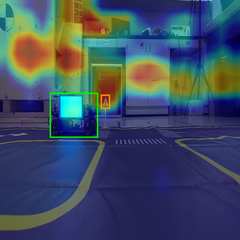}\hspace{0.0001\linewidth}
\includegraphics[width=0.19\linewidth]{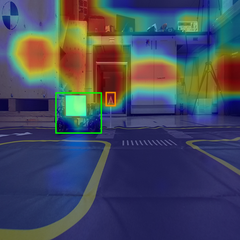}\hspace{0.0001\linewidth}
\includegraphics[width=0.19\linewidth]{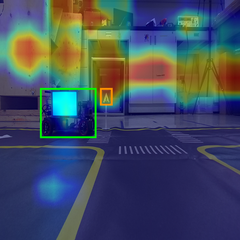}\hspace{0.0001\linewidth}

\caption{The position of objects in an image influences the performance of adversarial patches. The influence, as visualized in heatmaps, shifts according to changes in objects' positions. 
The color of a heatmap reflects the influence level (red to purple denoting most to least influence).}
\label{fig:heatmap}
\end{figure}
\section{Approach}
This section elaborates on our proposed approach for generating dynamic adversarial patches. Section~\ref{sec: dyn-env} explains the method for collecting patch training data that captures the dynamics of the environment. In Section~\ref{sec: sitnet}, we demonstrate the image transformation of the patch that is displayed on the screen. Finally, we describe the patch optimization in Section~\ref{sec: patch opt}.

\subsection{Dynamic Environment Settings}\label{sec: dyn-env}
The output of the object detection models (e.g., the regular yolov5s~\cite{yolov5}) such as cars or traffic signs can be greatly influenced by the objects' position in the image. This positioning directly impacts the model's focus or ``attention", altering the detection outcomes based on how far or near these objects are positioned relative to the camera's perspective.
To visualize the principal components of the learned features and representations, we apply Eigen-CAM~\cite{muhammad2020eigen,jacobgilpytorchcam} which reflects the influence level of areas of an image on the model through a heatmap (red to purple denoting most to least influence). As depicted in Figure~\ref{fig:heatmap}, the model's attention shifts according to changes in object positions. This in turn impacts the effectiveness of adversarial patches that are not overlaid on the target object. To compensate, a dynamic patch that adapts to the target's position could be a more effective approach. Without direct access to the \texttt{camera car}'s captured images during an attack, the \texttt{patch car} can estimate the target's position and adjust its patch accordingly to account for positional changes.\\

\begin{figure}[ht]
  \centering
  \begin{subfigure}{0.46\linewidth}
    \includegraphics[width=\linewidth]{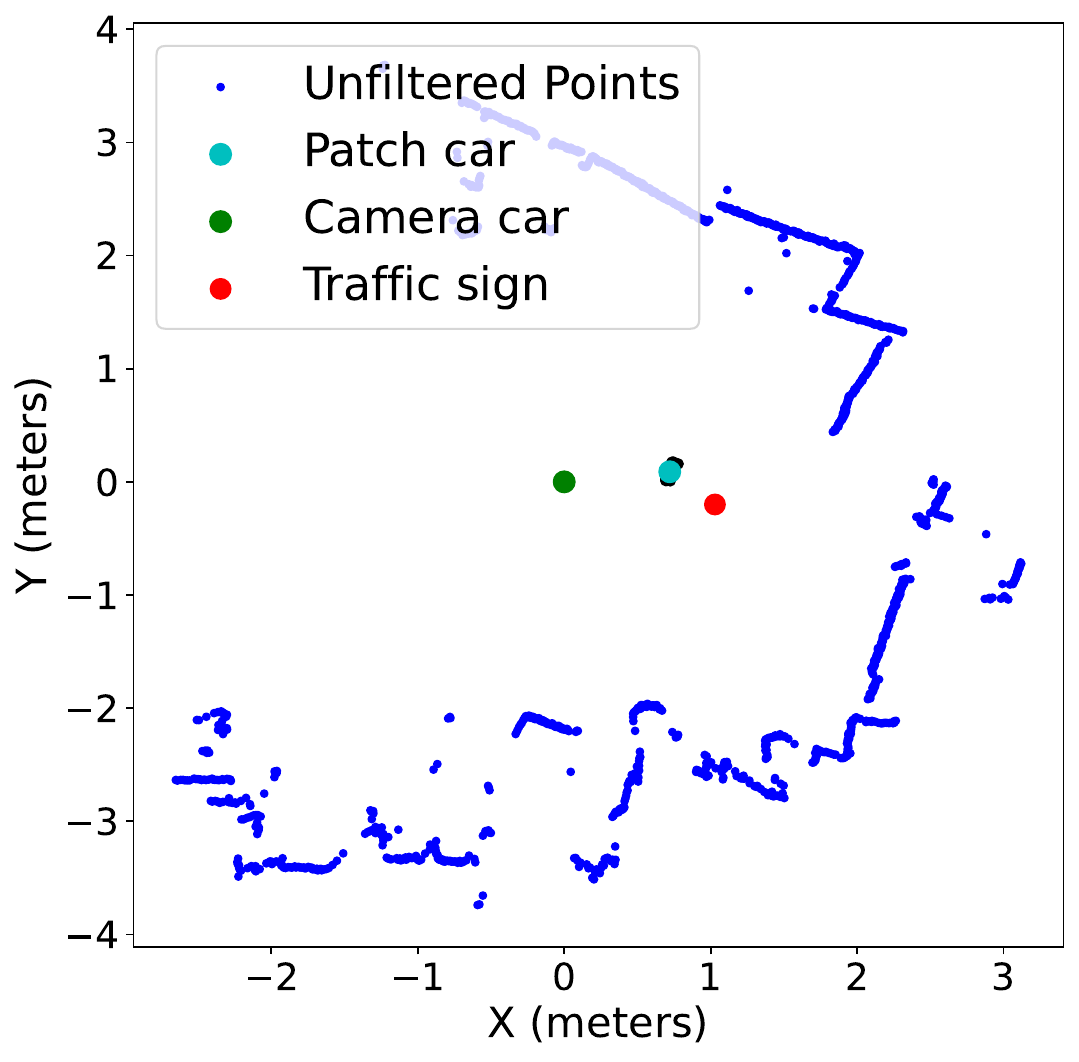}
    \caption{The coordinates of the cars and the target in one data frame where the \texttt{camera car}'s position is the origin.}
    \label{fig:lidar-map}
  \end{subfigure} ~
  \begin{subfigure}{0.485\linewidth}
    \includegraphics[width=\linewidth]{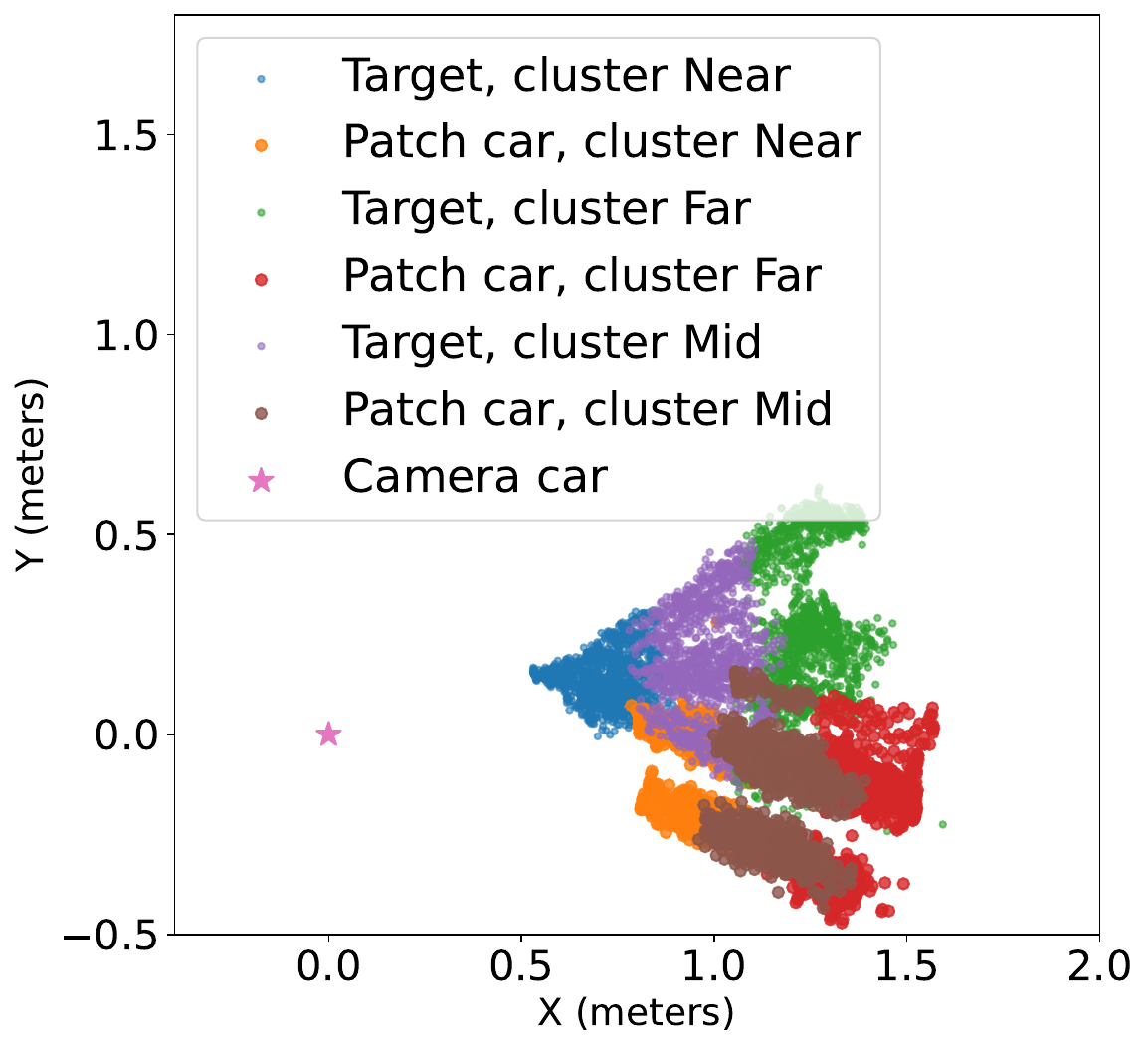}
    \caption{Clustered 2D position points of the \texttt{patch car} and the target where the \texttt{camera car}'s position is the origin.}
    \label{fig:cluster-plot}
  \end{subfigure}
  \caption{Analysis of data clustering based on positions: (a) Lidar map showing the coordinates of the cars and the target in a single frame; (b) Cluster plot of the \texttt{patch car} and the target for all frames in a dataset.}
  \label{fig:agents-formation}
\end{figure}

\begin{figure}[ht]
  \centering
  \begin{subfigure}[b]{0.9\linewidth}
  \setlength{\lineskip}{4pt}
    \centering 
    \includegraphics[width=0.2\linewidth]{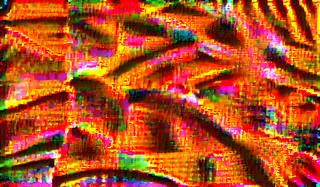}
    \includegraphics[width=0.2\linewidth]{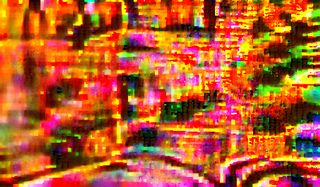}
    \includegraphics[width=0.2\linewidth]{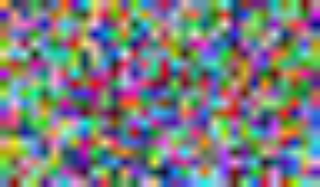}
    \includegraphics[width=0.2\linewidth]{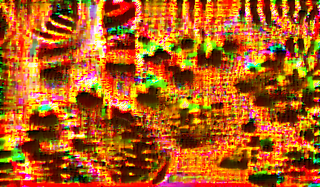}
    \\
    \includegraphics[width=0.2\linewidth]{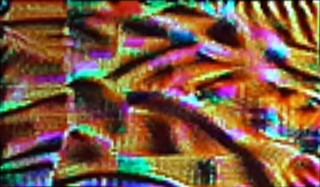}
    \includegraphics[width=0.2\linewidth]{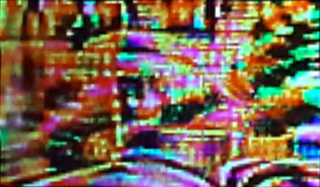}
    \includegraphics[width=0.2\linewidth]{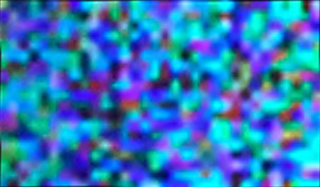}
    \includegraphics[width=0.2\linewidth]{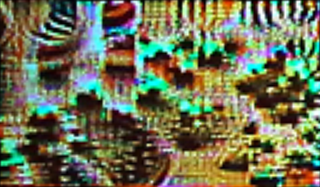}
    \\
    \includegraphics[width=0.2\linewidth]{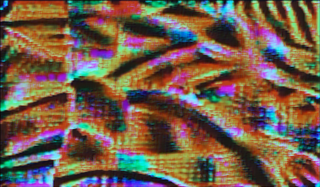}
    \includegraphics[width=0.2\linewidth]{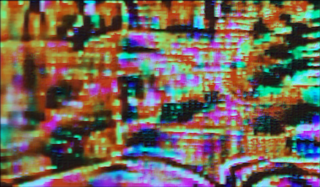}
    \includegraphics[width=0.2\linewidth]{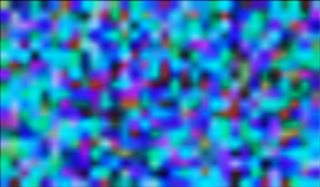}
    \includegraphics[width=0.2\linewidth]{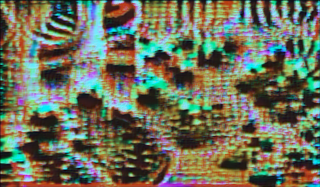}
    \caption{SIT-Net predictions. The first row shows input images (X), the second row shows camera-captured images (Y), and the third row shows the model's predictions.}
    \label{fig:sit-net}
  \end{subfigure}
  \begin{subfigure}[b]{0.8\linewidth}
    \centering
    \includegraphics[width=\linewidth]{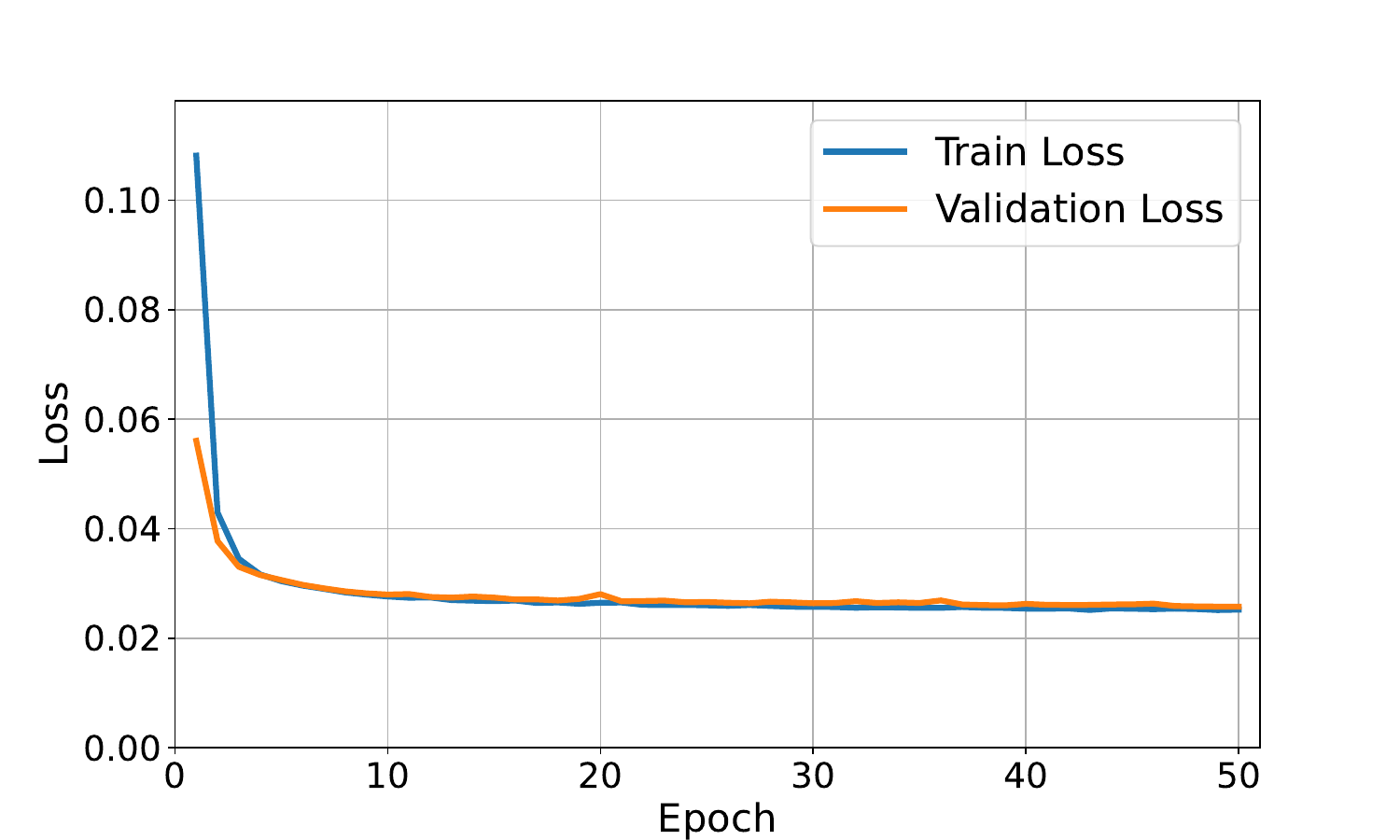}
    \caption{Loss over training epochs for SIT-Net.}
    \label{fig:sitnet-plot}
  \end{subfigure}
  \caption{SIT-Net predictions and the loss during its training.}
  \label{fig:combined sitnet}
\end{figure}

\subsubsection{Data collection}
We employ Lidar to obtain the positions of the cars and the target object.  Multiple autonomous driving scenarios involving the \texttt{patch car} and the target object are collected synchronously using Lidar and camera data. 
 
\noindent\textbf{Image and Lidar recording.} Both the \texttt{patch car} and the target object are recorded using a camera, and their positions are recorded via Lidar. This synchronized dual recording setup is designed to capture a variety of scenarios, including those where either one or both subjects are in motion. We enhance our position tracking accuracy using Lidar-based SLAM techniques, specifically AMCL~\cite{ros_amcl}, which affords robust pose estimation even in dynamic and unpredictable environments. Figure~\ref{fig:lidar-map} demonstrates the \texttt{camera car}'s map generated with lidar points.\\
\noindent\textbf{Screen contour identification.} A full blue image is displayed on the screen during the recording sessions. This practice assists in the accurate identification of screen contours, which is crucial for the perspective transformation of the patches during the training phase.\\
\subsubsection{Data clustering}
\label{sec:data-clustering}
The recorded data is then clustered by the K-means clustering algorithm based on the distance of the \texttt{patch car} and target object from the \texttt{camera car}.
This way, the positions of the \texttt{patch car} and the target in the images in the same cluster are within the same threshold. 
Each cluster represents a specific range of distances, and a dedicated patch is trained for each cluster. Figure~\ref{fig:cluster-plot} shows the clustered points of the \texttt{patch car} and the target. This enhances the performance of the patches that are optimized for different \rev{clusters}. \rev{Appendix~\ref{sec:Appendix-dataset} provides a more detailed overview of our dataset.}\\
\noindent\textbf{Dynamic patch display.} The trained patches are then employed in a dynamic fashion. The system utilizes the \texttt{camera car}’s position to determine the appropriate patch to display, ensuring a more effective and contextually relevant visual transformation.\\
This structured and strategic approach to data collection and processing ensures that our system is capable of handling a variety of real-world scenarios, making the patches dynamic and enhancing the overall effectiveness of the visual transformations.
\begin{figure*}[htp]
\centering
    \includegraphics[width=0.9\linewidth]{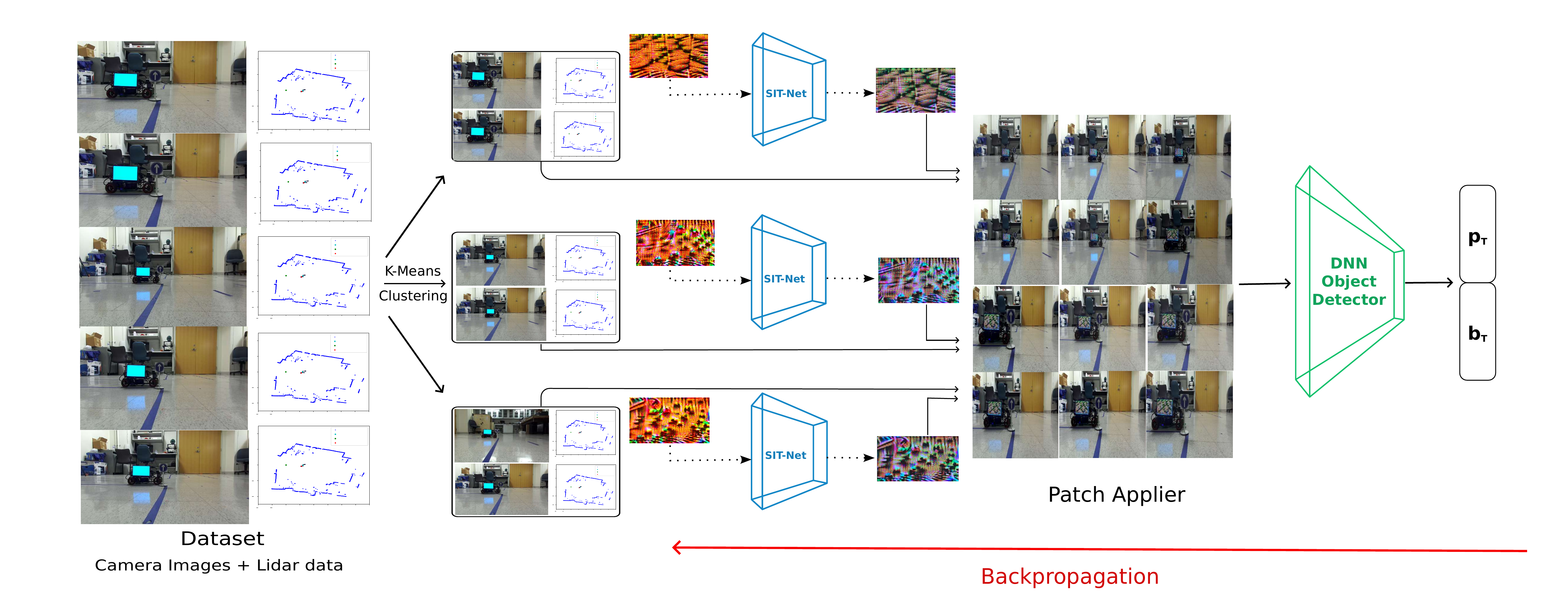}
    \caption{The patch optimization pipeline.}
    \label{fig: diagram}
\end{figure*}

\subsection{Screen Image Transformation Network (SITNet)}\label{sec: sitnet}

To model the visual changes of the patch in the images captured by the camera, we use a convolutional neural network (CNN) model. The SITNet is a CNN designed to transform the colors and contrast of images to mitigate the difference between simulation and the real world. It specifically focuses on adapting the appearance of images displayed on a screen, as captured by the camera, ensuring a more accurate representation and smoother transition between the simulated and physical environments. SITNet comprises two convolutional layers. In the following, we describe data processing and loss functions in the SIT-Net training steps. 
\subsubsection{Screen Data Collection and Prepossessing}
We manually collected the dataset used to train SITNet to address the challenges of discrepancies between simulated and real-world visual data. The collection procedure is as follows: 

\noindent\textbf{Input image generation.} A diverse set of patch-like images is created, ensuring a variety of color and contrast scenarios. These patches are of the same size as the targeted screen area to maintain consistency.

\noindent\textbf{Screen display.} The generated images are displayed on a screen, capturing a wide range of total variations to adequately represent different potential real-world conditions.

\noindent\textbf{Image recording.} A camera is utilized to record the displayed images. Precise coordination of the four corner points of the screen area is maintained to aid in the subsequent image-processing steps.

\noindent\textbf{Image preprocessing.} The captured images are subjected to cropping and perspective transformation based on the known screen coordinates. This step ensures that the images are accurately aligned and standardized, mirroring the conditions of the screen display.

This comprehensive data collection and preprocessing procedure ensures that the network is exposed to a wide spectrum of visual variations, enhancing its ability to generalize and accurately transform images in practical scenarios. 

\subsubsection{Loss Function and Optimization}
The mean squared error is used in the loss function to reduce the difference between predicted and real-world images. However, these pixel-level differences might not correlate well with how images are perceived or their similarity. Perceptual loss addresses this limitation by incorporating features extracted from pre-trained neural networks that are designed to mimic human visual perception. Moreover, a total variation term~\cite{tvloss} is added to the loss function to smooth the predicted images. 
\noindent\textbf{Mean squared error (MSE) loss.} We quantify the pixel-wise difference between the predicted and real-world images by MSE loss. 
\begin{equation}
    \gL_{\text{MSE}}(\hat{Y}, Y) = \frac{1}{N} \sum_{i=1}^{N} (\hat{Y}_i - Y_i)^2
\end{equation}
where \( N \) is the total number of pixels in the image.

\noindent\textbf{Perceptual loss.} The perceptual loss measures the difference in style between images at the feature level rather than the pixel level. The first nine layers of the pre-trained VGG19 network~\cite{Simonyan2014} are used to extract feature maps from the input and target images. Then the MSE between these feature maps is computed. This way, the model is forced to generate images that are perceptually similar to the target images. The perceptual loss is defined below. 
\begin{equation}
    \gL_{P}(\hat{Y}, Y) = \frac{1}{M} \sum_{j=1}^{M} (V(\hat{Y})_j - V(Y)_j)^2,
\end{equation}
where \( M \) is the total number of elements in the VGG feature maps.

\noindent\textbf{Total variation (TV) loss~\cite{tvloss}:} The TV loss below encourages spatial smoothness in the predicted image.
\begin{equation}
    \gL_{\text{TV}}(\hat{Y}, Y) = \sum_{i,j} \left| \hat{Y}_{i,j+1} - \hat{Y}_{i,j} \right| + \left| \hat{Y}_{i+1,j} - \hat{Y}_{i,j} \right|.
\end{equation}
The overall loss is a weighted sum of MSE loss, Perceptual loss, and TV loss.
\begin{equation}
    \gL(\hat{Y}, Y) = \gL_{\text{MSE}}(\hat{Y}, Y) + \alpha \gL_{P}(\hat{Y}, Y) + \beta \Delta \gL_{\text{TV}}(\hat{Y}, Y).
\end{equation}
The network is trained over $n_{\text{epochs}}=50$ epochs, with a learning rate of 0.001. The weights for the perceptual and total variation losses are set as \(\alpha=0.02\) and \(\beta=0.01\), respectively.
Figure~\ref{fig:combined sitnet} depicts the training curves and the SIT-Net predictions for an environment instance.

\begin{algorithm}[H]
\caption{Adversarial Patch Optimization with SITNet}
\label{alg:patch-train}
\begin{algorithmic}[1]
\REQUIRE Set of clustered images $\sX_{\text{clustered}}$, original traffic sign bounding box $\vb_{\text{orig}}$, patch $z \in \sP'$, learning rate $\eta$, number of iterations $N$, top-$k$ value $k$, SITNet model $M_{\text{SITNet}}$
\ENSURE Set of optimized patch $\sP^*$
\STATE Initialize $z$
\FOR {$x$ in $\sX_{\text{clustered}}$}
    \FOR {$i = 1$ to $N$}
        \STATE $z' \gets M_{\text{SITNet}}(z)$
        \STATE $\pi(x,s_c,s_p) = x + \delta(s_{c},s_{p}, \phi)$
        \STATE $Y=f(\pi(x, s_c, s_p), \theta)$
        \STATE $\sM \gets \text{Filter}(Y, \vb_{\text{orig}})$
        \label{alg:line: filter}
        \STATE $\sM_k \gets \text{Top-k}(\sM)$
        \STATE $\text{IoU}_k \gets \text{IoU}(\sM_k, \vb_{\text{orig}})$
        \label{alg:line: iou}
        \STATE $\mathcal{O} \gets \frac{1}{k} \sum_{j=1}^{k} \left( \text{Conf}_{\sM_k[j]} \times \text{IoU}_{k[j]} \right)$
        \STATE $z \gets z + \eta \cdot \nabla_z \mathcal{O}$
    \ENDFOR
    \STATE $\sP^* \leftarrow z$
\ENDFOR

\RETURN $\sP^*$
\end{algorithmic}
\end{algorithm}
\subsection{Patch Optimization}\label{sec: patch opt}
The optimization procedure for the patch is described in Algorithm~\ref{alg:patch-train}.
We optimize a patch for each cluster in the dataset. First, the patch is transformed by the SIT-Net. Then we apply the transformed patch, $z'$, to the images and pass it to the object detector. The line~\ref{alg:line: filter} includes filtering of the object detector's prediction and extraction of all bounding boxes in the area of the original object. The objectiveness probabilities of these bounding boxes are integrated into the class probabilities. Then we extract the top-$k$ target class confidence values.
Subsequently, in line~\ref{alg:line: iou}, the IoU of the top-$k$ target class bounding boxes with the original object bounding box is computed. In the next step, the IoU values are multiplied by the top-$k$ confidence, and its mean is calculated. This value represents the alignment and confidence of the selected bounding boxes. Finally, this value is utilized in the optimization process of the patch. Figure~\ref{fig: diagram} shows a pipeline of the patch optimization process. In summary, this process begins with the clustering of the images through the K-Means method using LiDAR data, then the patches that are transformed by the SIT-Net are applied to images of each cluster. The patches are optimized by backpropagating. 

\section{Evaluation}
We test our proposed dynamic attack in different scenarios where two robotic agents interact with each other and must follow traffic signs.
\subsection{Experimental Setup}
\noindent\textbf{Agents.} We use two Rosmaster R2 robot cars with Ackermann structure~\cite{yahboom_ros} equipped with Lidar and IMU sensors for localization and mapping. The \texttt{camera car} uses a Zed2 camera~\cite{stereolabs_zed_2} for image recording. The \texttt{patch car} is equipped with a seven-inch LCD screen attached to its side to display the patches. The car's decision-making planner employs object detection results to plan for its motions. Object detection and patch training are carried out by a server using a NVIDIA RTX4090 GPU.
\\
\noindent\textbf{Object detector model.} For each robot car, we fine-tuned the model from yolov5~\cite{yolov5} to detect traffic signs and the other robot car. The model trained over 1.8 K images with the weights from a pre-trained model yolov5s and the input image size of $640 \times 640$. Our customized dataset is attached.\footnote{The link to dataset is provided within supplementary materials.}

\begin{figure}[h]
  \centering
        \includegraphics[height=2.1cm]{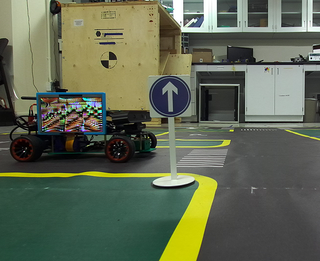}
        \includegraphics[height=2.1cm]{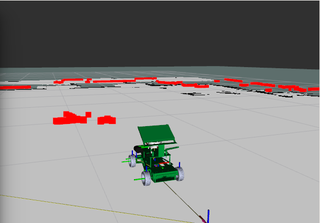}
        \includegraphics[height=2.1cm]{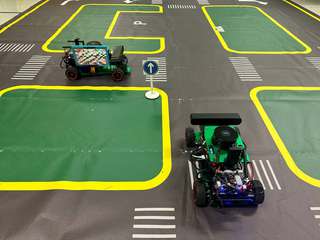}
  \caption{Experimental setup. Left column: images recorded by the \texttt{camera car}; Middle column: point clouds representing the \textit{patch car} and the target sign within the map; Right column: the top view of the two cars.}
  \label{fig:combined slam}
\end{figure}

\begin{figure*}[h]
\centering
 \begin{subfigure}{0.3\linewidth}
    \includegraphics[width=\linewidth]{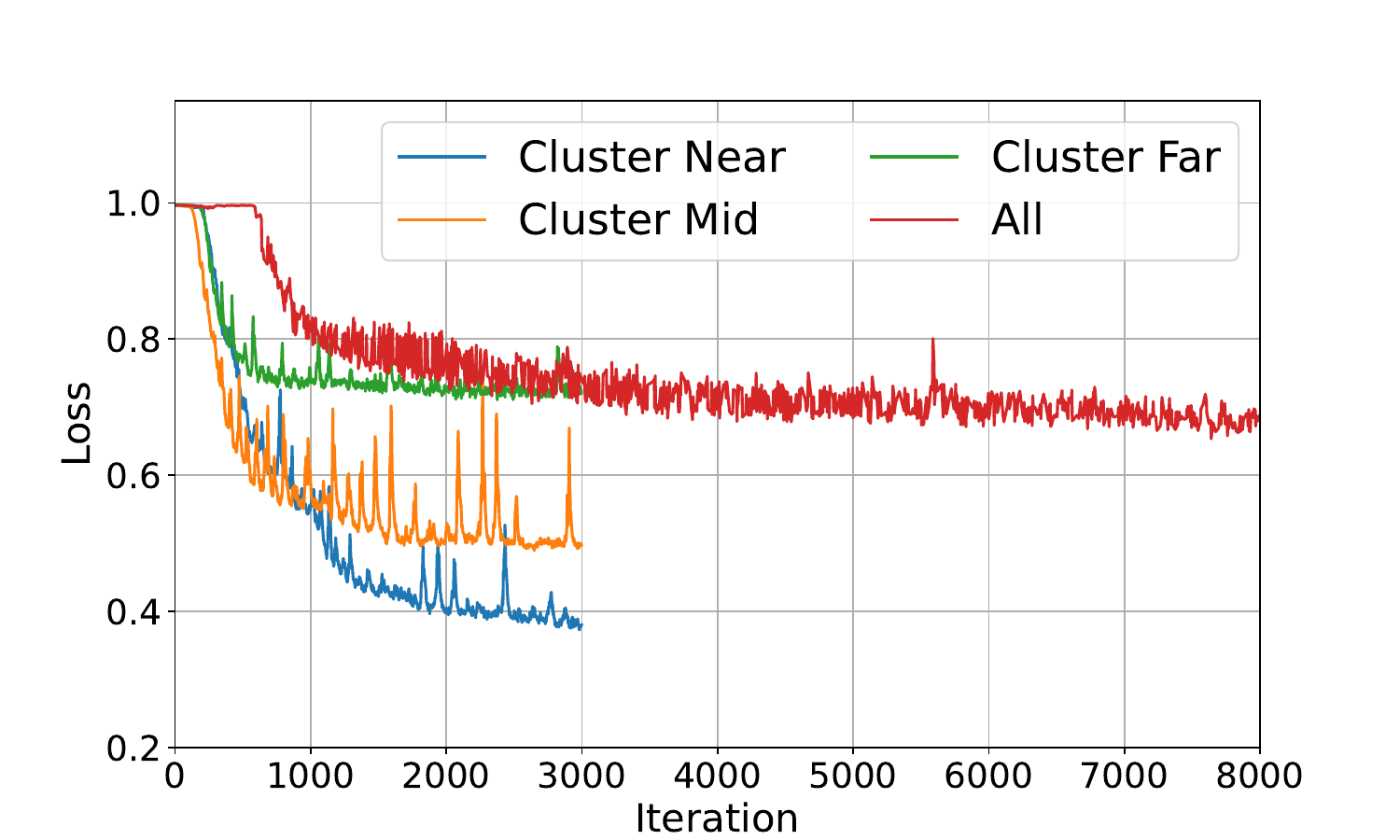} 
    \caption{Loss for the ``Go-straight" sign attack}
    \end{subfigure}
\begin{subfigure}{0.3\linewidth}
    \includegraphics[width=\linewidth]{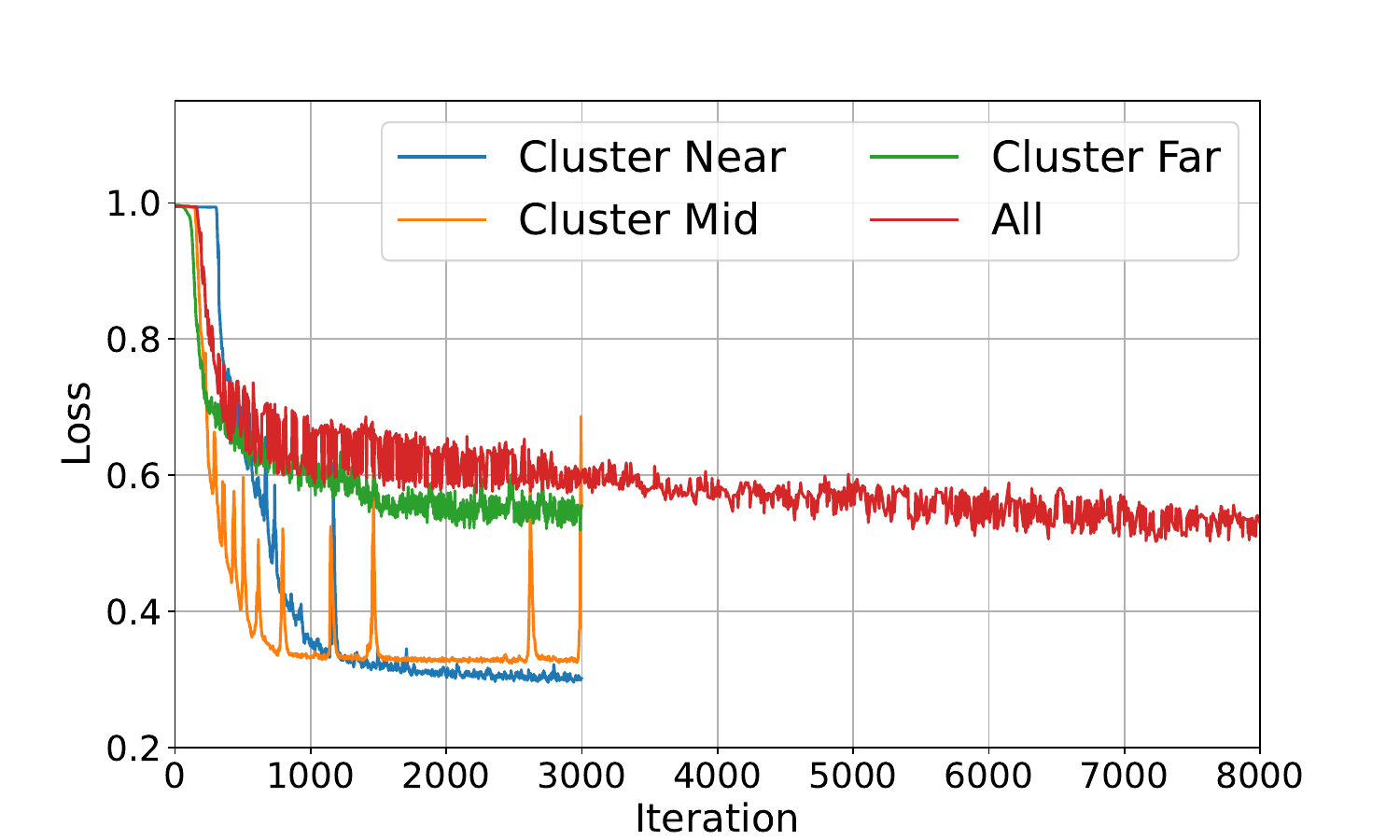} 
    \caption{Loss for the ``Turn" sign attack}
    \end{subfigure}
\begin{subfigure}{0.3\linewidth}
    \includegraphics[width=\linewidth]{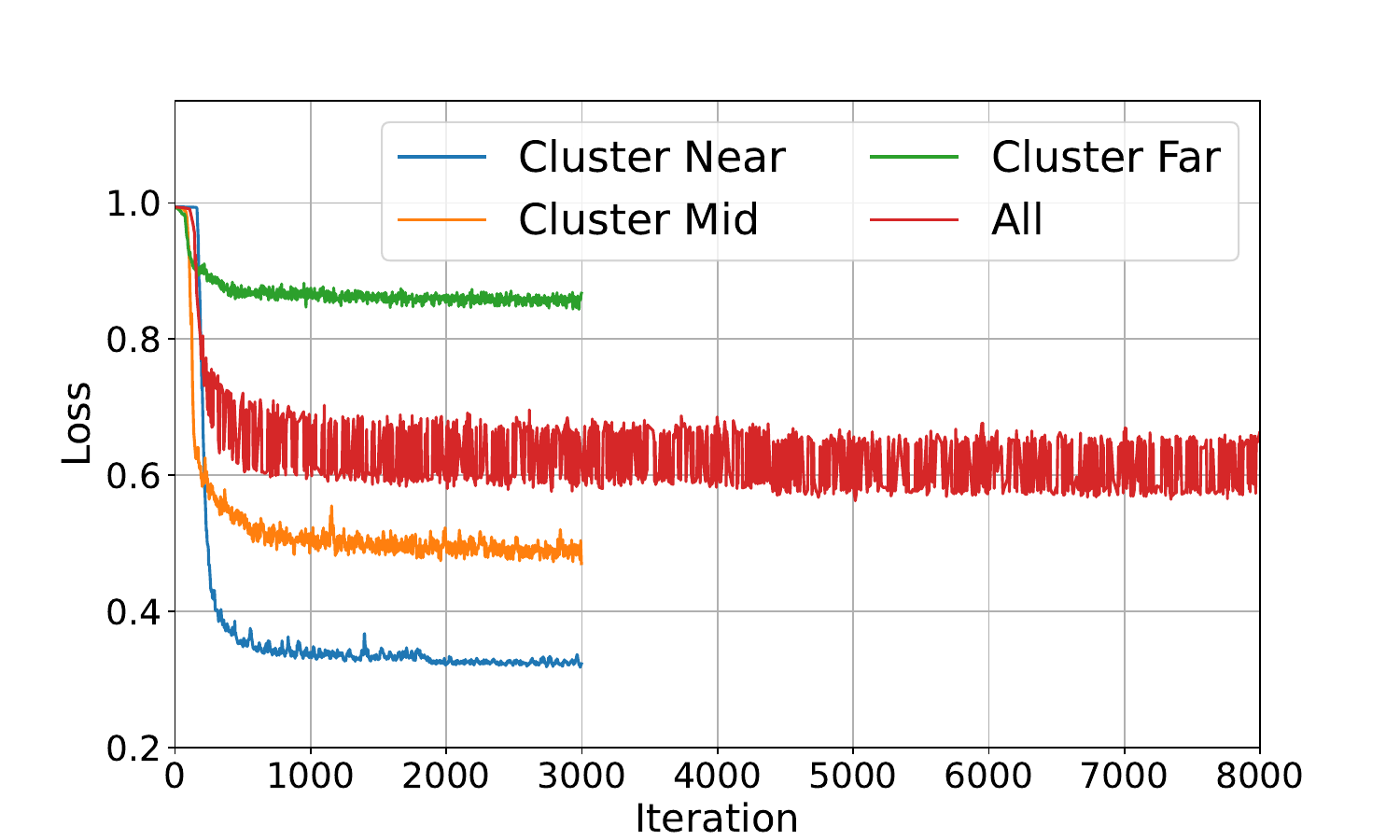} 
    \caption{Loss for the ``Pedestrian" sign attack}
    \end{subfigure}
    \caption{Patch training loss for different traffic sign attacks, showing the rapid convergence of training for individual clusters compared to the entire dataset.}
    \label{fig:loss-over-iteration}
\end{figure*}

\noindent\textbf{Autonomous driving scenarios.} We consider various robot positional settings at the intersections where the traffic signs are crucial for decision-making. When the \texttt{patch car} must yield to the \texttt{camera car} according to the traffic signs, it aims to alter the class of the detected signs by the \texttt{camera car} to pass the intersection earlier. In this setting, both cars approach an intersection within the same time window while driving on crossed roads, and the \texttt{patch car} tries to pass the intersection before the \texttt{camera car}. There are three possible traffic signs: Go-straight, Turn, and Pedestrian crossing signs. The sign is targeted by the patch attack to be misclassified as a stop sign. Figure~\ref{fig:combined slam} shows our setup for the experiment.

\subsection{Attack Procedure}

\noindent\textbf{Step 1.} We collect images from the display screen to construct the SIT-Net as explained in Section~\ref{sec: sitnet}.

\noindent\textbf{Step 2.} We run robot cars in each scenario to collect synchronized Lidar and image data. This dataset is clustered as explained in Section~\ref{sec: dyn-env}. The number of clusters hinges on the total distance traversed and the precision of positional estimations.  For our experiments, we set the cluster count to three. We utilize this processed dataset to optimize a set of patches using Algorithm~\ref{alg:patch-train}.  We set the training epoch at 1000 for each cluster and start with a randomly generated initial patch. Concurrently, a static patch is trained across all clusters for an identical epoch count.

\noindent\textbf{Step 3.}  After training, we evaluate the effectiveness of the generated patches by the attack success rate in an extensive number of experiments including a similar environment and background and in settings with a background that is not in the attack dataset. For each of the sign types, approximately 1000 image frames are evaluated. An attack is considered successful if the patch makes the object detector predict a stop sign with confidence greater than 0.5 in place of the original sign.

\rev{We expand our evaluation to include comparisons with methodologies similar to ours, particularly focusing on sticker-based attacks as explained in~\cite{Thys2019}, which represent the closest parallel in existing literature. Unlike previous studies where patches are directly attached to the target object, in our approach, the adversarial patch and the target object are positioned separately. This unique setup necessitated the optimization of our patches through Equation~\ref{eq: objective} coupled with the consideration of the non-printability score, which is elaborated in~\cite{Thys2019, Sharif2016}. 

Moreover, we conduct experiments in which no patch was displayed to have a benign case.
Table~\ref{table:attack-success-rate} presents the results of these experiments. 
}

\begin{table}[H]
  \centering
  \begin{tabular}{@{}lcccc@{}}
    \toprule
    \textbf{Attack Method} & \textbf{Dynamic (\%)} & \textbf{Static (\%)} &
    \rev{\textbf{Printed (\%)}} & \rev{\textbf{Benign (\%)}}\\
    \midrule
    Go-straight sign & 39.9 & 30.0 & \rev{1.3} & \rev{0.0}\\
    Turn sign & 27.4 & 22.1 & \rev{3.0} & \rev{0.0}\\
    Pedestrian sign & 18.0 & 15.2 & \rev{2.0} & \rev{0.0}\\
    \bottomrule
  \end{tabular}
  \caption{\rev{Comparison of attack success rates for Dynamic, Static, and Printed Patches across different sign types in an intersection scenario. This table also shows the control group (``Benign") where no adversarial attack was present, indicating no false detections under benign conditions.}}
  \label{table:attack-success-rate}
\end{table}

\begin{figure}[h]
    \centering
    \begin{subfigure}{0.24\linewidth}
    \setlength{\lineskip}{4pt}
    \centering
        \includegraphics[width=0.9\linewidth]{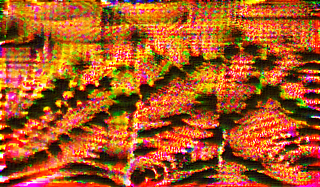}
        \\
        \includegraphics[width=0.9\linewidth]{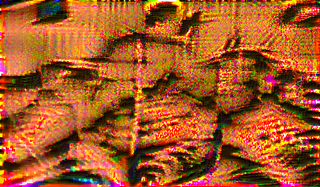}
        \\
        \includegraphics[width=0.9\linewidth]{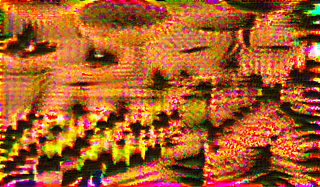}
    \end{subfigure}
    \begin{subfigure}{0.24\linewidth}
    \setlength{\lineskip}{4pt}
    \centering
        \includegraphics[width=0.9\linewidth]{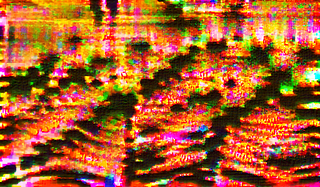}
        \\
        \includegraphics[width=0.9\linewidth]{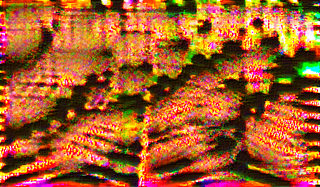}
        \\
        \includegraphics[width=0.9\linewidth]{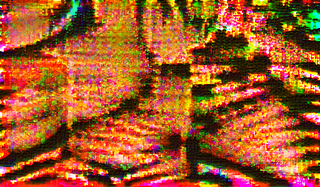}
    \end{subfigure}
    \begin{subfigure}{0.24\linewidth}
    \setlength{\lineskip}{4pt}
    \centering
        \includegraphics[width=0.9\linewidth]{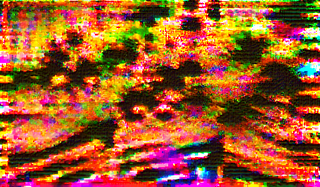}
        \\
        \includegraphics[width=0.9\linewidth]{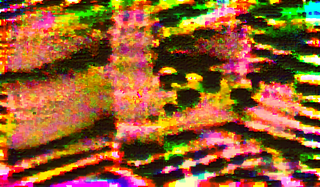}
        \\
        \includegraphics[width=0.9\linewidth]{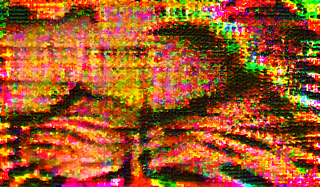}
    \end{subfigure}
    \begin{subfigure}{0.24\linewidth}
    \setlength{\lineskip}{4pt}
    \centering
        \includegraphics[width=0.9\linewidth]{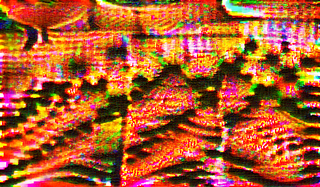}
        \\
        \includegraphics[width=0.9\linewidth]{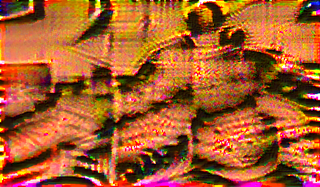}
        \\
        \includegraphics[width=0.9\linewidth]{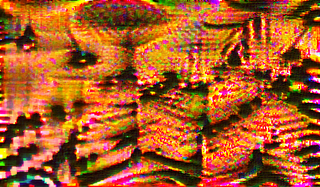}
    \end{subfigure}
    \caption{Display of trained adversarial patches across various clusters and the entire dataset. The rows represent different traffic sign scenarios, specifically ``Go-straight", ``Turn", and ``Pedestrian" signs, in that order. For the columns, from left to right, we showcase patches optimized for clusters based on proximity to the target: ``Near", ``Mid", and ``Far", followed by patches trained on the combined dataset (denoted as `All'). This arrangement illustrates how the patches are tailored to specific distances from the target object.}
    \label{fig:patch-gallery}
\end{figure}

\begin{figure*}[]
\centering
\begin{subfigure}{0.32\linewidth}
\includegraphics[width=\linewidth]{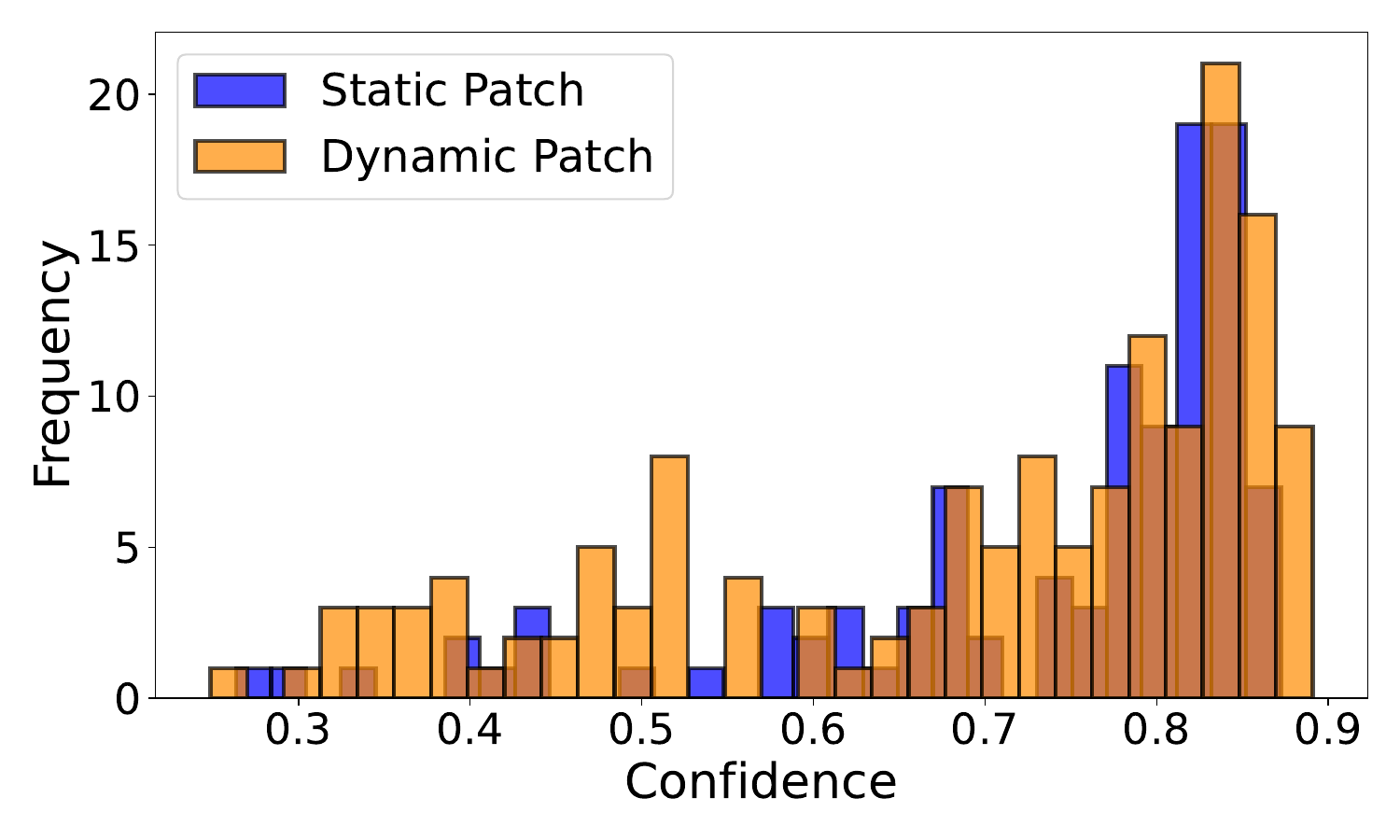}
\caption{\rev{Confidence distribution for misclassifying "Go-straight" sign as "Stop-sign"}}
\end{subfigure}
\hfill
\begin{subfigure}{0.32\linewidth}
\includegraphics[width=\linewidth]{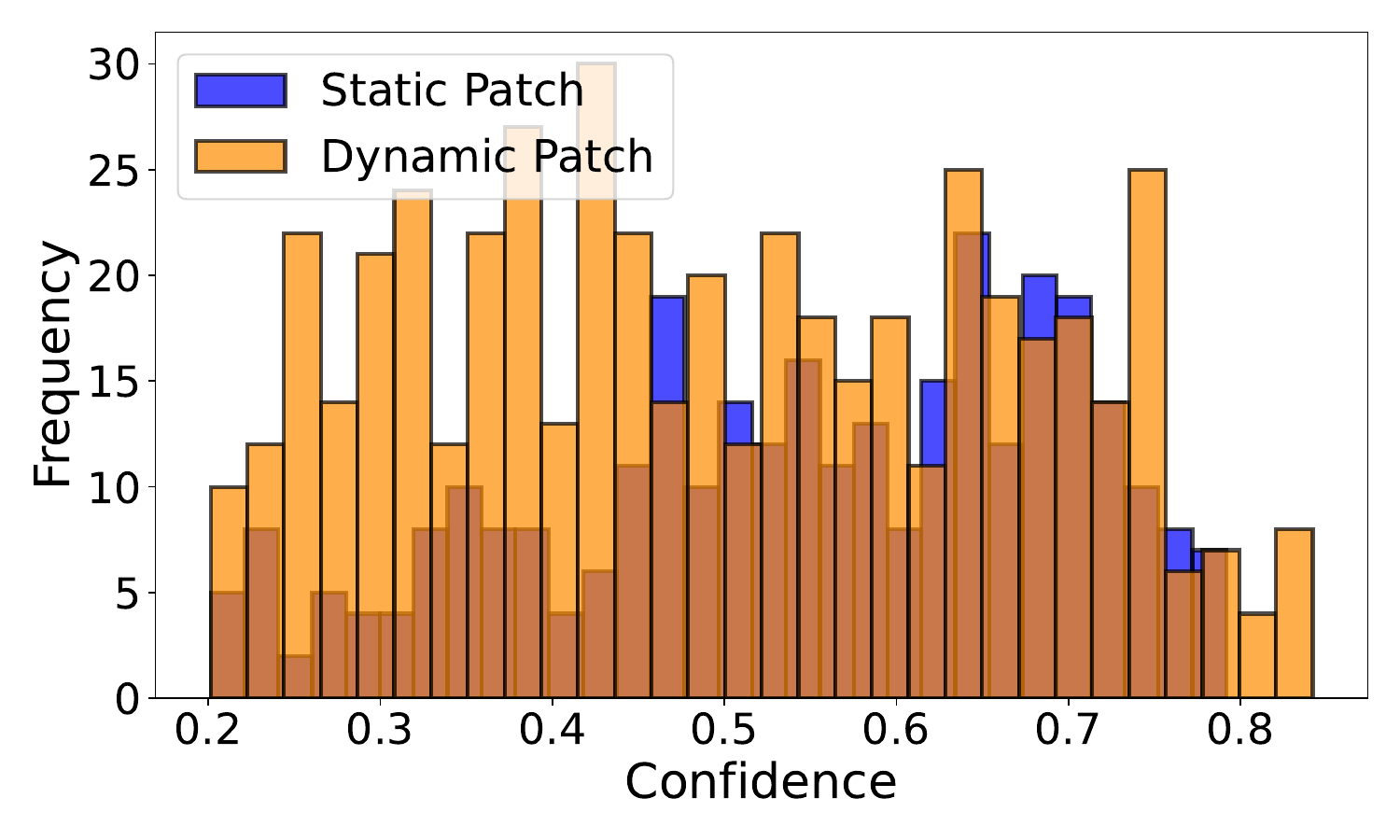}
\caption{\rev{Confidence distribution for misclassifying "Turn" sign as "Stop-sign"}}
\end{subfigure}
\hfill
\begin{subfigure}{0.32\linewidth}
\includegraphics[width=\linewidth]{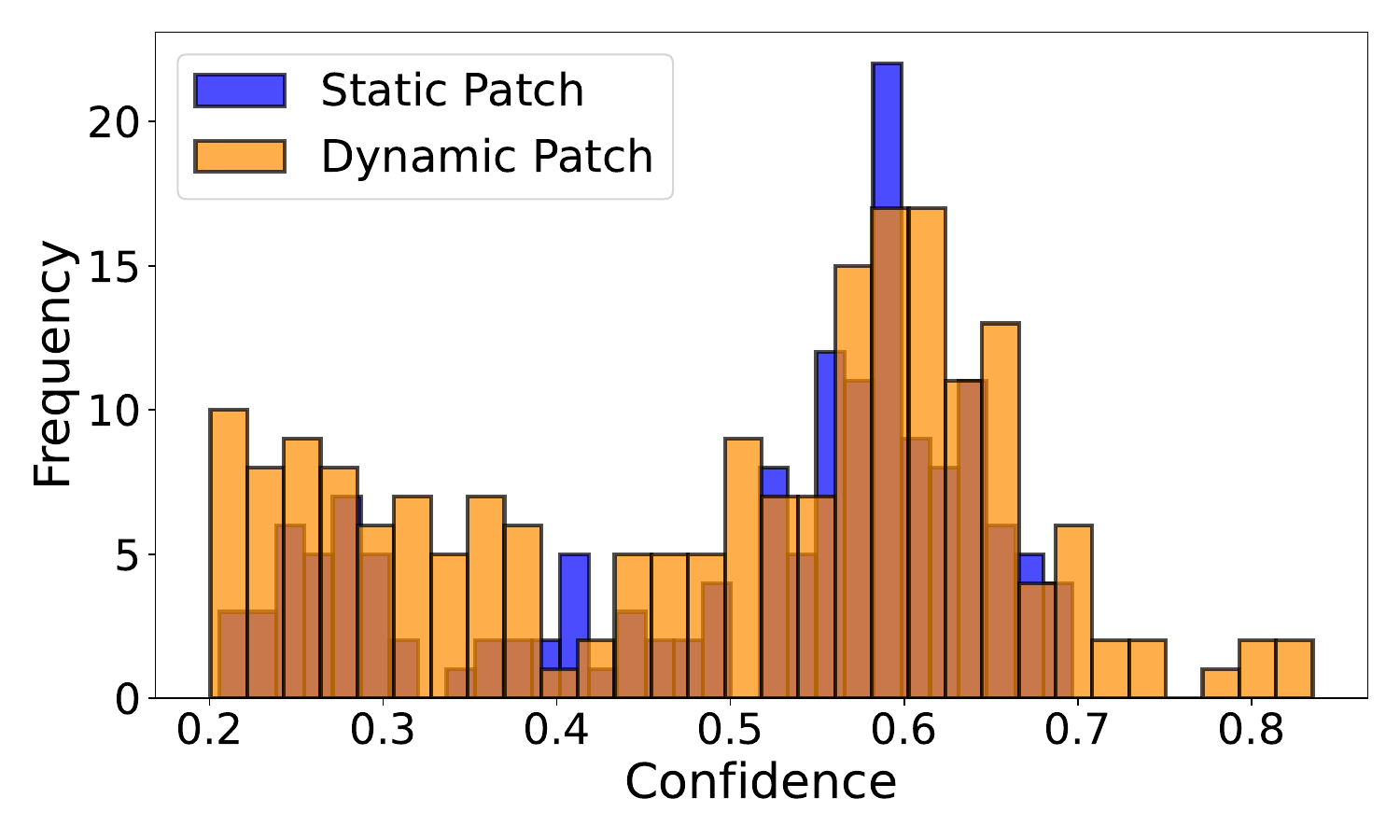}
\caption{\rev{Confidence distribution for misclassifying "Pedestrian" sign as "Stop-sign"}}
\end{subfigure}

\begin{subfigure}{0.32\linewidth}
\includegraphics[width=\linewidth]{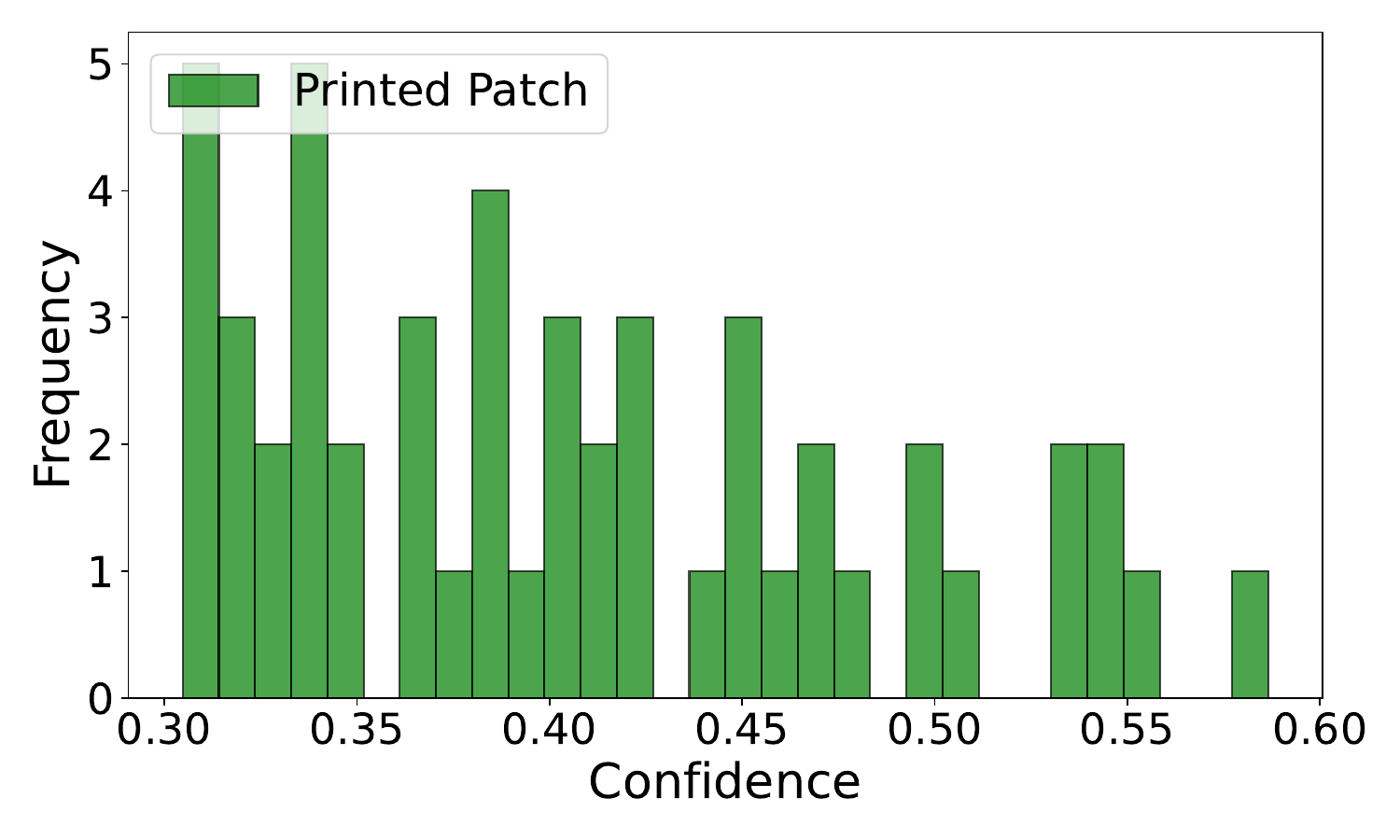}
\caption{\rev{Confidence distribution for misclassifying "Go-straight" sign as "Stop-sign" using a printed patch}}
\end{subfigure}
\hfill
\begin{subfigure}{0.32\linewidth}
\includegraphics[width=\linewidth]{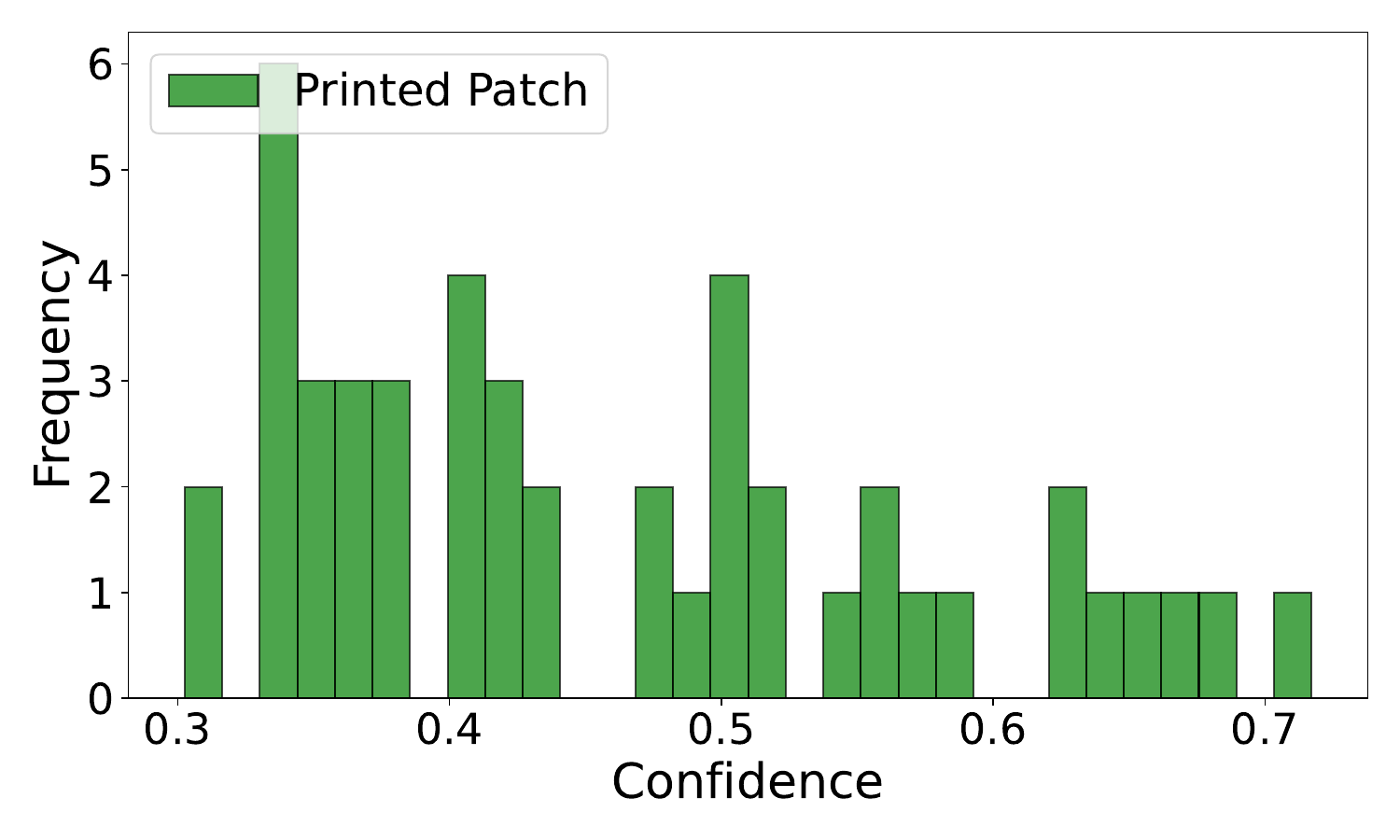}
\caption{\rev{Confidence distribution for misclassifying "Turn" sign as "Stop-sign" using a printed patch}}
\end{subfigure}
\hfill
\begin{subfigure}{0.32\linewidth}
\includegraphics[width=\linewidth]{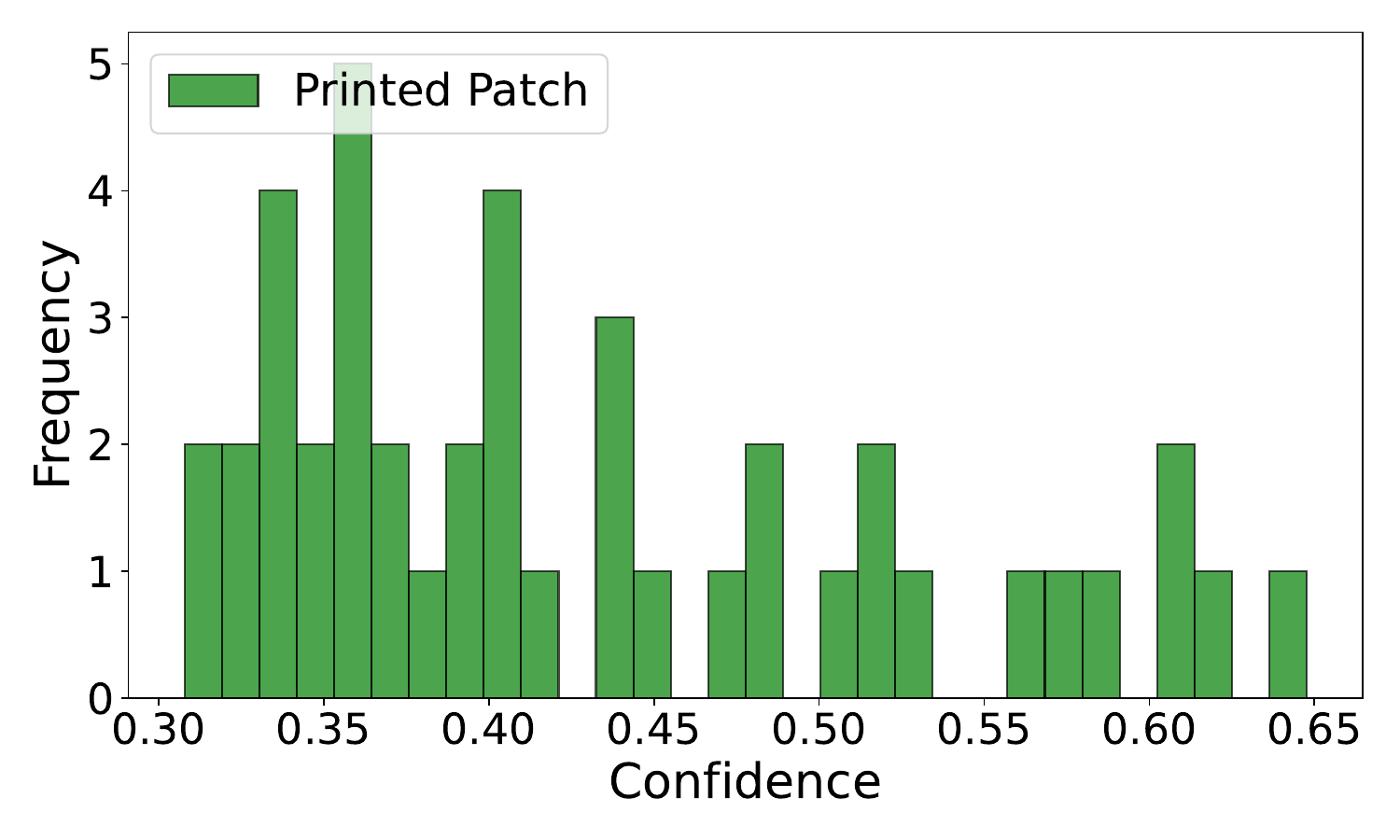}
\caption{\rev{Confidence distribution for misclassifying "Pedestrian" sign as "Stop-sign" using a printed patch}}
\end{subfigure}
\caption{\rev{Comparative histograms illustrating the confidence levels in misclassifying "Go-straight", "Turn", and "Pedestrian" signs as "Stop-sign" under two conditions: with dynamic/static patches (top row) and with printed adversarial patches (bottom row). These graphs highlight the effectiveness and variation of the adversarial attacks across different scenarios, noting that higher confidence values indicate better performance in the attacks. The spread and peaks of the confidence levels evidence the effectiveness of dynamic patches in deceiving.}}
\label{fig:Conf-Dist}
\end{figure*}

\begin{figure*}[]
\centering
\begin{subfigure}{0.32\linewidth}
    \centering
    \adjustbox{valign=t}{\includegraphics[width=\linewidth]{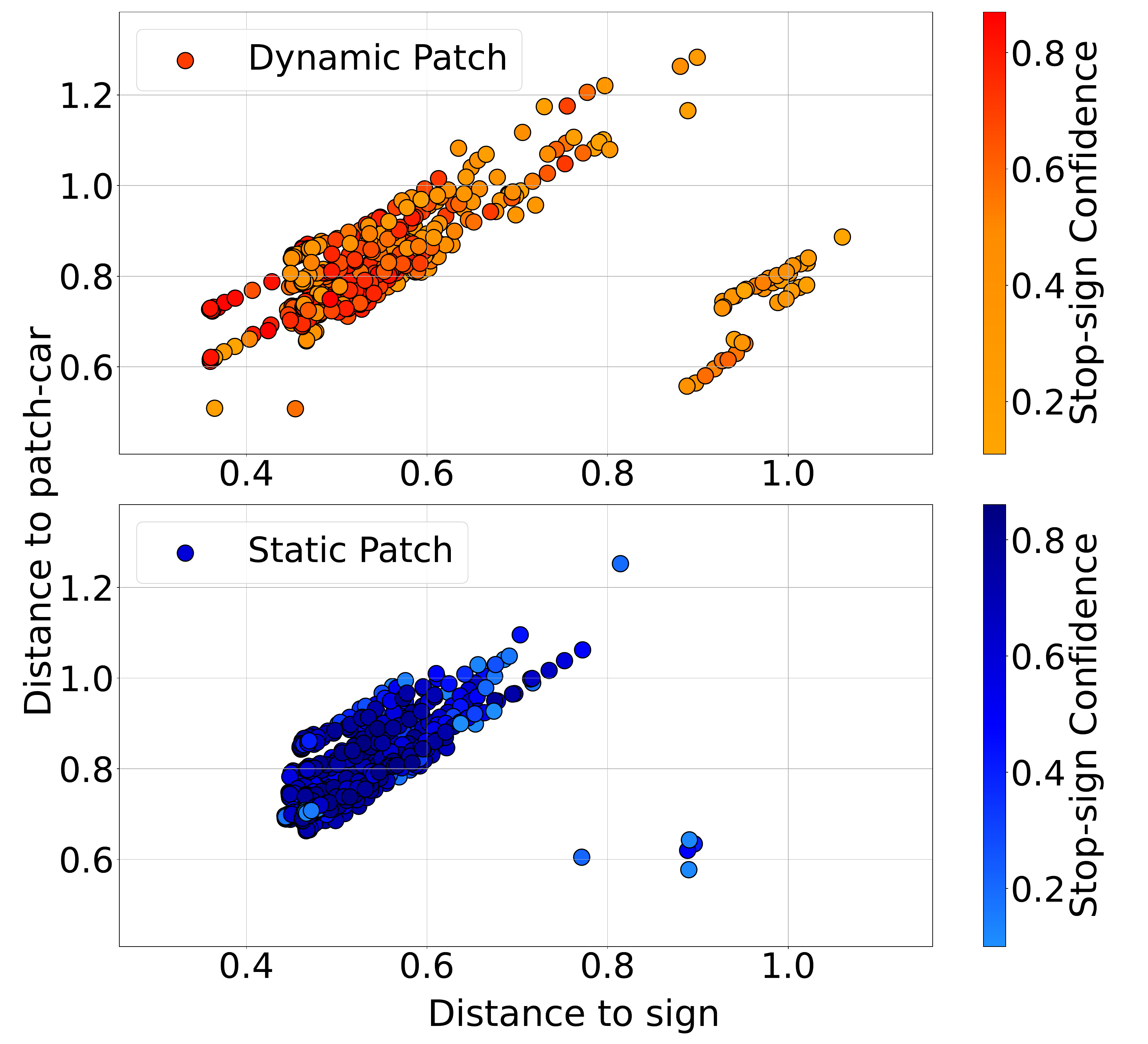}}
    \caption{\rev{Effectiveness range of patches for the "Go-straight" sign attack}}
\end{subfigure}\hfill
\begin{subfigure}{0.32\linewidth}
    \centering
    \adjustbox{valign=t}{\includegraphics[width=\linewidth]{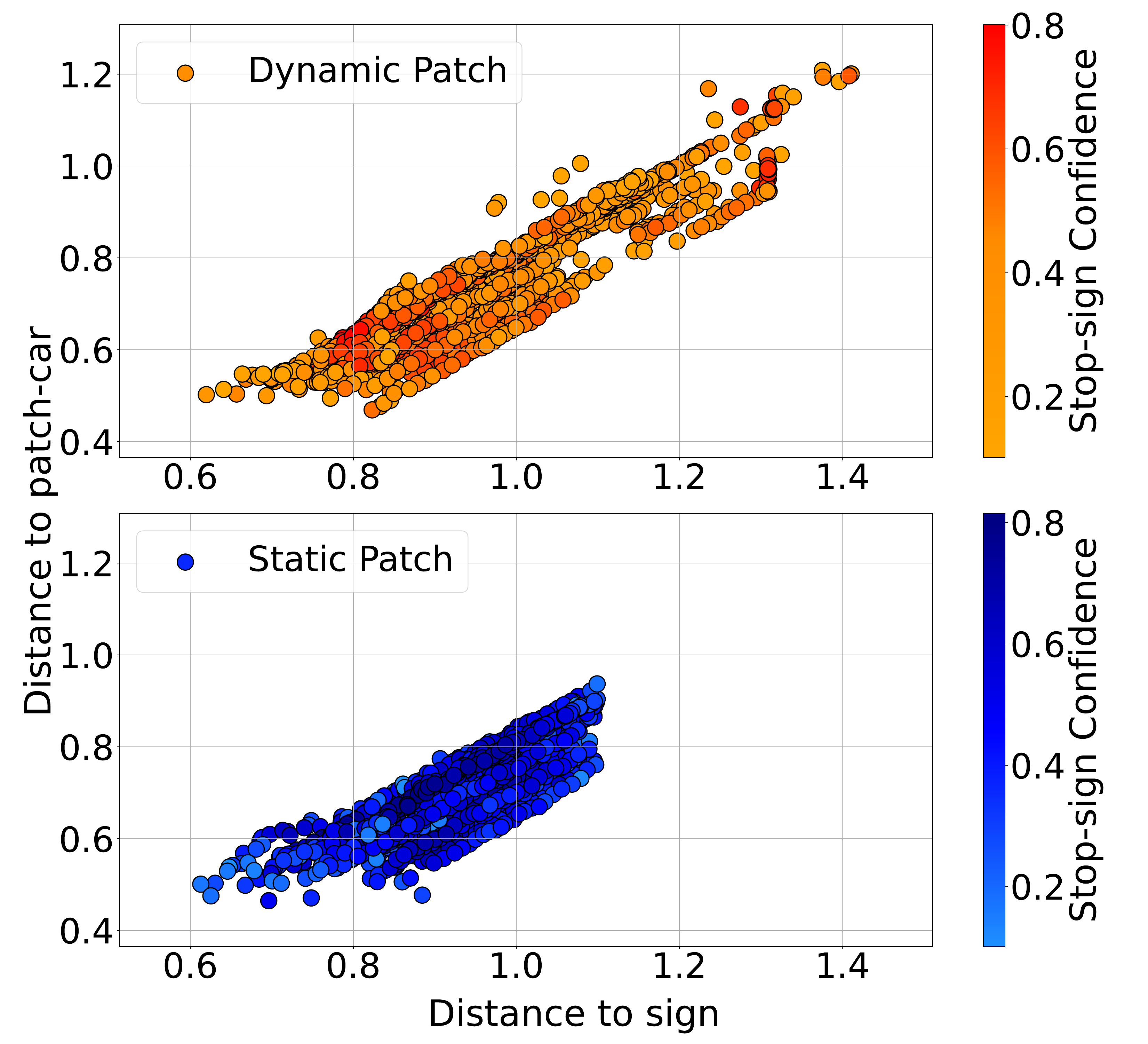}}
    \caption{\rev{Effectiveness range of patches for the "Turn" sign attack}}
\end{subfigure}\hfill
\begin{subfigure}{0.32\linewidth}
    \centering
    \adjustbox{valign=t}{\includegraphics[width=\linewidth]{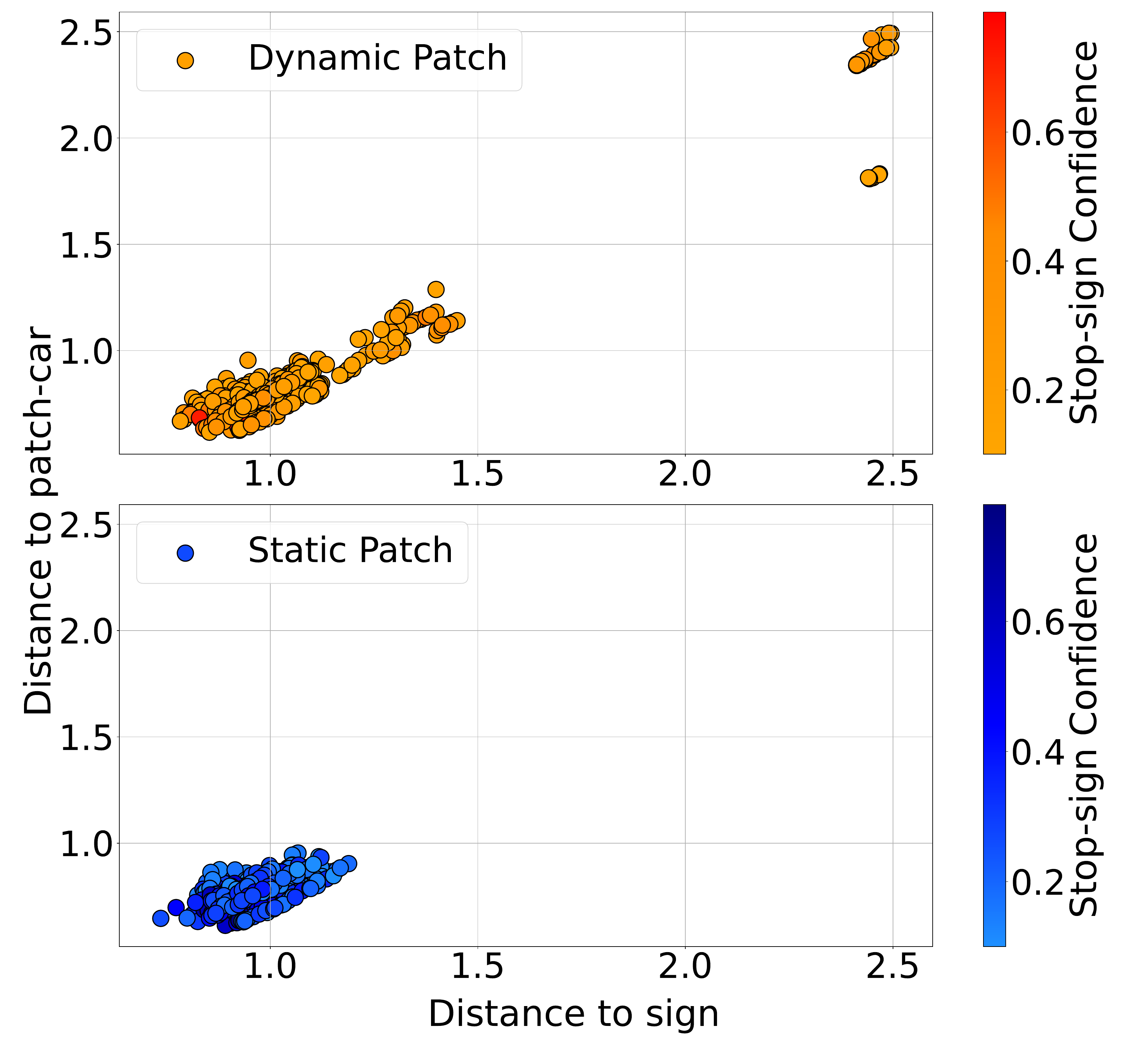}}
    \caption{\rev{Effectiveness range of patches for the "Pedestrian" sign attack}}
\end{subfigure}
\caption{Scatter plots contrasting the effectiveness of dynamic and static patches in ``Stop-sign" adversarial attacks as a function of the distance (in meters) of \texttt{camera car} to both the sign and the \texttt{patch car}. The dynamic patches (displayed in red-orange) demonstrate higher effectiveness across a more extensive range of distances from the camera to the patch, as indicated by the spread of the data points. The color intensity reflects the classifier's confidence in misclassification, with the dynamic patch showing high confidence over larger distances compared to the static patch (shown in blue).}
\label{fig:Range}
\end{figure*}


\subsection{Results and Discussion}

\begin{figure*}[htp]
\centering
\begin{subfigure}{\textwidth}
  \centering
  \setlength{\lineskip}{4pt}
    
  \includegraphics[width=0.24\linewidth]{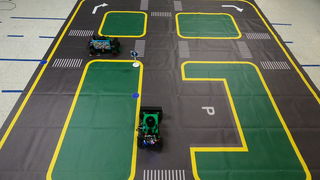}\hspace{0.0001\linewidth}
  \includegraphics[width=0.24\linewidth]{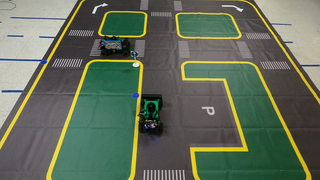}\hspace{0.0001\linewidth}
  \includegraphics[width=0.24\linewidth]{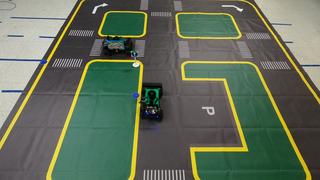}\hspace{0.0001\linewidth}
  \includegraphics[width=0.24\linewidth]{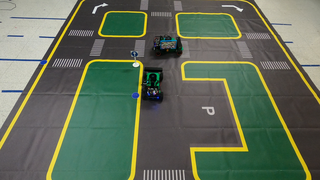}\\
  \includegraphics[width=0.24\linewidth]{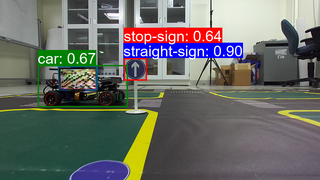}\hspace{0.0001\linewidth}
  \includegraphics[width=0.24\linewidth]{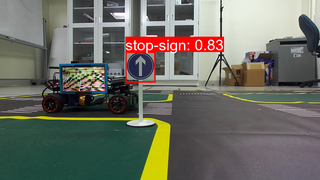}\hspace{0.0001\linewidth}
  \includegraphics[width=0.24\linewidth]{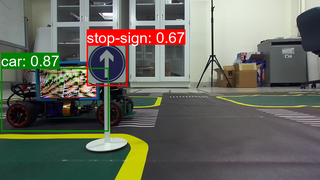}\hspace{0.0001\linewidth}
  \includegraphics[width=0.24\linewidth]{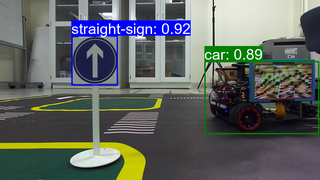}\\
  \includegraphics[width=0.24\linewidth]{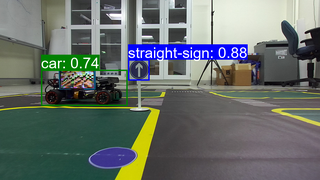}\hspace{0.0001\linewidth}
  \includegraphics[width=0.24\linewidth]{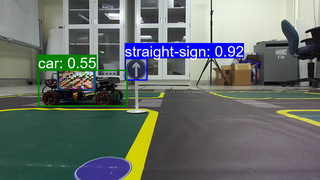}\hspace{0.0001\linewidth}
  \includegraphics[width=0.24\linewidth]{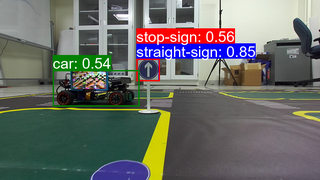}\hspace{0.0001\linewidth}
  \includegraphics[width=0.24\linewidth]{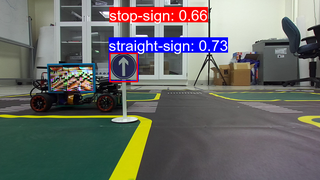}
  \caption{Adversarial patch attacks on the Go-straight sign.}
\end{subfigure}

\vspace{1em} 

\begin{subfigure}{\textwidth}
  \centering
  \setlength{\lineskip}{4pt}

  \includegraphics[width=0.24\linewidth]{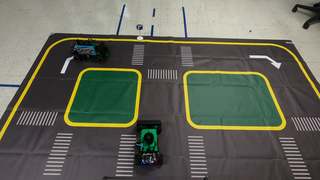}\hspace{0.0001\linewidth}
  \includegraphics[width=0.24\linewidth]{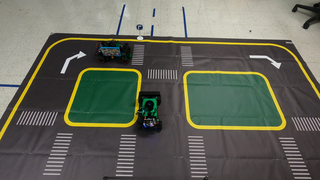}\hspace{0.0001\linewidth}
  \includegraphics[width=0.24\linewidth]{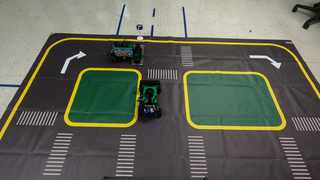}\hspace{0.0001\linewidth}
  \includegraphics[width=0.24\linewidth]{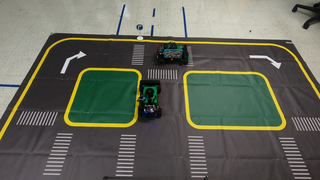}\\
  \includegraphics[width=0.24\linewidth]{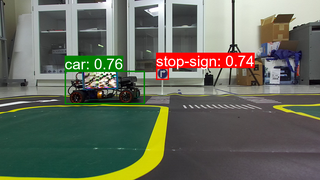}\hspace{0.0001\linewidth}
  \includegraphics[width=0.24\linewidth]{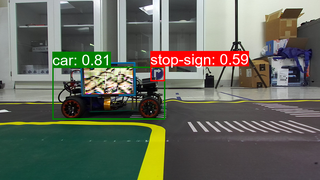}\hspace{0.0001\linewidth}
  \includegraphics[width=0.24\linewidth]{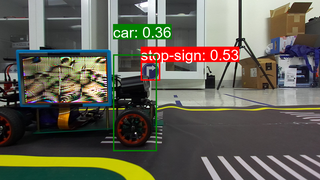}\hspace{0.0001\linewidth}
  \includegraphics[width=0.24\linewidth]{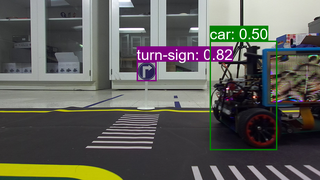}\\
  \includegraphics[width=0.24\linewidth]{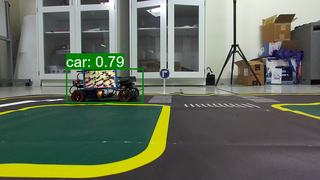}\hspace{0.0001\linewidth}
  \includegraphics[width=0.24\linewidth]{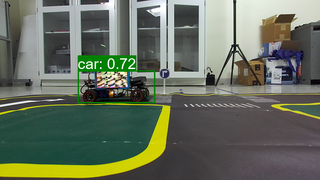}\hspace{0.0001\linewidth}
  \includegraphics[width=0.24\linewidth]{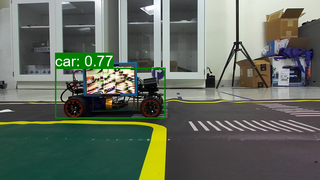}\hspace{0.0001\linewidth}
  \includegraphics[width=0.24\linewidth]{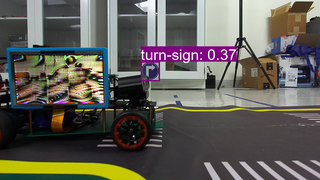}
  \caption{Adversarial patch attacks on the Turn sign.}
\end{subfigure}

\caption{Sequential frames depicting two types of adversarial patch attacks on an autonomous vehicle's perception system. For each sub-figure, the top row shows the top view of a dynamic patch attack scenario. The middle row illustrates a dynamic patch attack, wherein the adversarial patches displayed on the screen change based on the distances to the camera and the sign, leading to a more deceptive misclassification across consecutive frames. The bottom row presents a static patch attack, where a single, unchanging patch is used. Both attacks aim to disrupt the perception system, causing it to mistakenly identify the signs as Stop signs.}
\label{fig:Experiments}
\end{figure*}

The loss over iterations during training of the patch for each cluster and all datasets is depicted in Figure~\ref{fig:loss-over-iteration}. Also, Figure~\ref{fig:patch-gallery} shows the optimized patches for different clusters and the entire dataset. Figure~\ref{fig:loss-over-iteration} demonstrates that optimizing adversarial patches specifically for individual data clusters significantly expedites the training process and enhances the overall effectiveness of the attack. This cluster-based approach benefits from the inherent homogeneity within each cluster, allowing for more rapid optimization compared to a normal training approach on a broader dataset. \rev{The success rate in Table~\ref{table:attack-success-rate}indicate that Dynamic patches consistently outperform Static and Printed patches that attempted to cover all clusters in the dataset. Printed patches show significantly lower effectiveness across all sign types. We discus the ineffectiveness of these printed patches in Appendix~\ref{sec:Appendix-printpatch}.  The benign case, with a 0.0\% success rate, confirms the absence of false positives in normal conditions.}
Further, Figure~\ref{fig:Conf-Dist} displays the confidence distribution for the misclassification of ``Stop-sign" for three sets of adversarial patch attacks on the ``Pedestrian", ``Turn", and ``Go-straight". The dynamic patch consistently outperforms the static patch in every case in terms of confidence level. A visual depiction of the attack efficiency corresponding to the \texttt{camera car}'s distance from the \texttt{patch car} and the sign is shown in Figure\ref{fig:Range}. In contrast to the static patch, which has a more concentrated cluster of effectiveness at shorter ranges, the dynamic patch's effectiveness does not decrease as significantly with greater distance. Overall, Figure~\ref{fig:Conf-Dist} and Figure~\ref{fig:Range} point to the fact that dynamic patches are more successful than static patches in deceiving the object detector over a variety of instances. This can be explained by the adaptive nature of dynamic patches, which are designed to take into account different perspectives and distances, making them more resilient to environmental changes.
This is visually represented in Figure~\ref{fig:Experiments}, which depicts the consecutive frames of the attacks. A video showing attacks using dynamic patches in various autonomous driving scenarios involving two R2 robot cars is attached. \footnote{ \href{https://youtu.be/Wh2sPYpWczQ}{https://youtu.be/Wh2sPYpWczQ}}

Moreover, the designed objective function demonstrates the efficiency in generating a stop sign in the location of the original traffic sign as shown in Figure~\ref{fig:Experiments}. This functionality is essential for the success of the adversarial attack, as it directly influences the decision-making algorithms of the autonomous driving systems being attacked. 
By incorrectly identifying key traffic signs, the dynamic patches deceive the autonomous vehicle into stopping or passing prematurely, leading to incorrect navigational choices.

\rev{\subsection{Ablation Study on the Effectiveness of SIT-Net}

To evaluate the necessity and impact of the Screen Image Transformation Network (SIT-Net) within our adversarial framework, we conducted an ablation study where both static and dynamic adversarial patches were trained without incorporating SIT-Net. This experimental setup allowed us to isolate and study the specific contributions of SIT-Net, which simulates realistic environmental influences like color transformation and contrast adjustments that are crucial in real-world scenarios.

In our simulations, we overlaid the trained patches on images to mimic their real-world deployment on screens. The patches performed effectively under these ideal conditions, demonstrating their potential efficacy. However, when these patches were actually displayed on screens in a real-world environment, their effectiveness drastically declined, with a success rate dropping to zero. This significant discrepancy underscores the critical role of SIT-Net in preparing adversarial patches to cope with real-world variability and environmental conditions.

Furthermore, we found that simulations lacking SIT-Net do not adequately capture the visual characteristics of adversarial patches as displayed in real-world conditions. In simulations, the patches may appear sharper and have clearer colors than they do in real-world environments. This difference results in an overestimation of patch effectiveness during optimization, as the simulated visual quality exceeds what is typically achievable in practical deployments. This emphasizes the essential role of SIT-Net in providing a realistic preparation and evaluation of adversarial patches, ensuring that their performance under real-world conditions is accurately represented.

Figure \ref{fig:NSC} showcases the dramatic changes in patches when displayed on a screen, highlighting the importance of including SIT-Net in the training process to achieve realistic and effective adversarial patches in practical deployments.

\begin{figure}[h]
    \centering
    \begin{subfigure}{0.32\linewidth}
    \setlength{\lineskip}{2pt}
    \centering
        \includegraphics[width=0.95\linewidth]{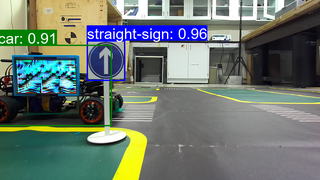}
        \\
        \includegraphics[width=0.95\linewidth]{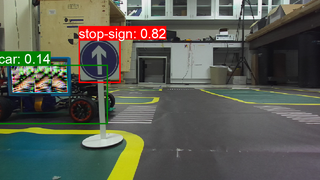}

    \end{subfigure}
    \begin{subfigure}{0.32\linewidth}
    \setlength{\lineskip}{2pt}
    \centering
        \includegraphics[width=0.95\linewidth]{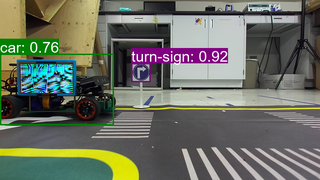}
        \\
        \includegraphics[width=0.95\linewidth]{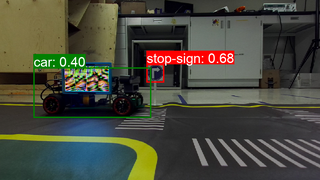}

    \end{subfigure}
    \begin{subfigure}{0.32\linewidth}
    \setlength{\lineskip}{2pt}
    \centering
        \includegraphics[width=0.95\linewidth]{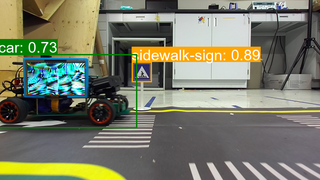}
        \\
        \includegraphics[width=0.95\linewidth]{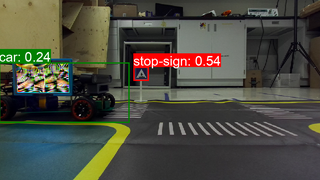}

    \end{subfigure}
    
    \caption{\rev{Illustration of adversarial patch effectiveness without SIT-Net. The top row depicts real-world implementation results, and the bottom row shows simulation results. The columns, from left to right, represent attacks on ``Go-straight", ``Turn", and ``Pedestrian" signs. This visual comparison clearly shows that the simulation without SIT-Net fails to replicate the real-world effectiveness of the patches, emphasizing the importance of SIT-Net in realistic adversarial training.}} 
    \label{fig:NSC}
\end{figure}


}

\subsection{Conclusions}
We formulate a novel approach for attacking object detectors by employing dynamic patches displayed on a screen to influence the decisions of autonomous driving systems. We have demonstrated that data clustering substantially accelerates the training process and enhances the overall efficacy of the dynamic attack. Additionally, to address the issue of color transformation in captured images of the displayed patch, we have introduced a straightforward CNN model capable of compensating for environmental factors. 

Our findings underscore the potential vulnerabilities in the perception mechanisms of autonomous driving systems and highlight the importance of considering such adversarial tactics in the development of robust decision-making algorithms. \rev{Future work will focus on developing strategies to enhance the robustness of these systems against adversarial attacks. This includes exploring advanced machine learning techniques to detect and mitigate the effects of dynamic adversarial patches in real-time, as well as enhancing the training methods to incorporate a wider range of environmental variations.}

\newpage
\bibliographystyle{plainnat}
\bibliography{references}

\begin{figure}[H]
    \centering
    \begin{subfigure}[b]{0.32\textwidth}
        \centering
        \includegraphics[width=\textwidth]{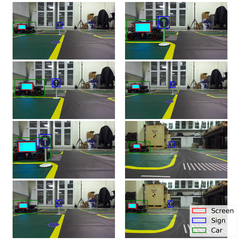}
        \caption{\rev{Go-straight sign}}
    \end{subfigure}
    \hfill
    \begin{subfigure}[b]{0.32\textwidth}
        \centering
        \includegraphics[width=\textwidth]{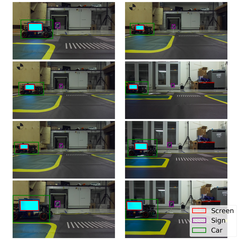}
        \caption{\rev{Turn sign}}
    \end{subfigure}
    \hfill
    \begin{subfigure}[b]{0.32\textwidth}
        \centering
        \includegraphics[width=\textwidth]{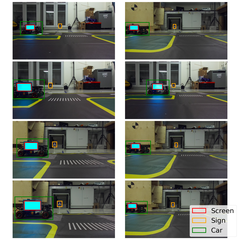}
        \caption{\rev{Pedestrian sign}}
    \end{subfigure}
    \caption{\rev{Example images from dataset for each sign, demonstrating the variety of scenarios in the dataset.}}
    \label{fig:dataset_examples}
\end{figure}

\onecolumn
\appendix
\rev{\section{Appendix}
\subsection{Dataset Overview}
\label{sec:Appendix-dataset}
The dataset utilized for training our adversarial patches was meticulously compiled to ensure a broad representation of various scenarios encountered in autonomous driving. This dataset includes synchronized LiDAR and image data collected from our robot cars in indoor environments for each sign, which is critical for training both static and dynamic adversarial patches effectively. Figure~\ref{fig:dataset_examples} displays examples from our dataset, showing the diversity in the images. The full dataset is provided in this \href{https://drive.google.com/drive/folders/1UiODhj44Wos0TJAiK1067lCwvnoJt0qu?usp=sharing}{link}\footnote{https://drive.google.com/drive/folders/1UiODhj44Wos0TJAiK1067lCwvnoJt0qu?usp=sharing}.}

\begin{figure}[H]
    \centering
    \begin{subfigure}[b]{0.3\textwidth}
        \centering
        \includegraphics[width=\textwidth]{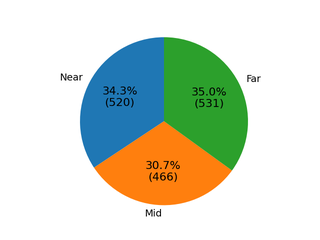}
        \caption{\rev{Go-straight sign}}
    \end{subfigure}
    \hfill
    \begin{subfigure}[b]{0.3\textwidth}
        \centering
        \includegraphics[width=\textwidth]{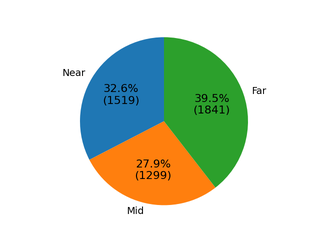}
        \caption{\rev{Turn sign}}
    \end{subfigure}
    \hfill
    \begin{subfigure}[b]{0.3\textwidth}
        \centering
        \includegraphics[width=\textwidth]{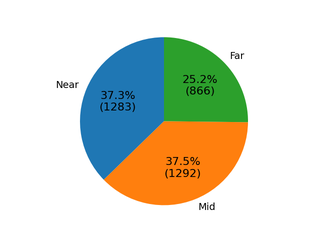}
        \caption{\rev{Pedestrian sign}}
    \end{subfigure}
    \caption{\rev{Distribution of images across the three clusters, showing the balance of clusters for each sign}.}
    \label{fig:dataset_distribution}
\end{figure}

\rev{The dataset for each sign is divided into three distinct clusters based on the distance to the screen and sign as discussed in section~\ref{sec:data-clustering}, which are pivotal for the effectiveness of dynamic patches in varying scenarios. These clusters are specifically designed to encapsulate different ranges of distances between the camera car and the target objects, ensuring that each patch is optimized for its respective range. The clustering approach helps in tailoring the patches to specific environmental conditions and object proximities. Figure \ref{fig:dataset_distribution} provides a detailed breakdown of the number of images in each cluster for each type of sign.}

\begin{figure}[H]
    \centering
    \setlength{\lineskip}{2pt}
    \includegraphics[width=0.3\linewidth]{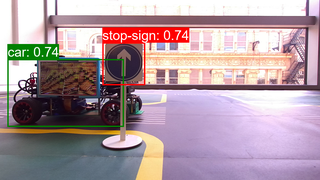}
    \includegraphics[width=0.3\linewidth]{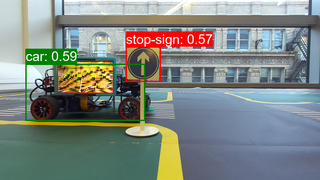}
    \includegraphics[width=0.3\linewidth]{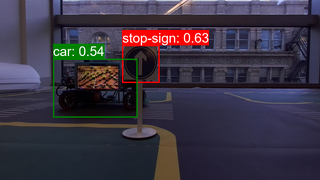}\\
    
    \includegraphics[width=0.3\linewidth]{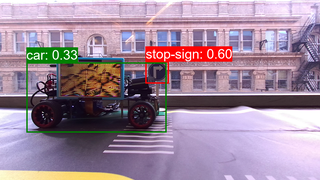}
    \includegraphics[width=0.3\linewidth]{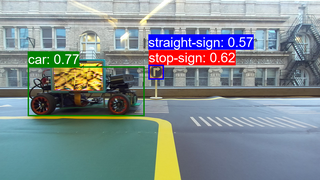}
    \includegraphics[width=0.3\linewidth]{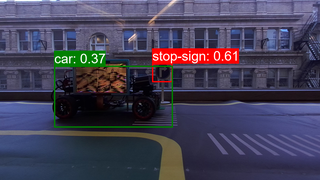}
    \caption{\rev{Adversarial patch attacks in natural lighting conditions at different times of the day.}}
    \label{fig:outdoor-test}
\end{figure}

\newpage
\rev{\subsection{Natural Lightening Condition Test}
We conducted additional experiments to assess the robustness of our adversarial patches under various outdoor lighting conditions using patches that were initially trained in an indoor environment. Even though our adversarial patches were trained for indoor controlled conditions, they remained effective in outdoor lighting conditions. Figure \ref{fig:outdoor-test} illustrates the performance of adversarial patches under different natural lighting conditions at various times of the day. These results highlight the resilience of our approach across a broad spectrum of natural lighting situations, though they also highlight areas for potential enhancement, particularly in very harsh lighting conditions. }

\begin{figure}[H]
    \centering
    \includegraphics[width=0.7
\linewidth]{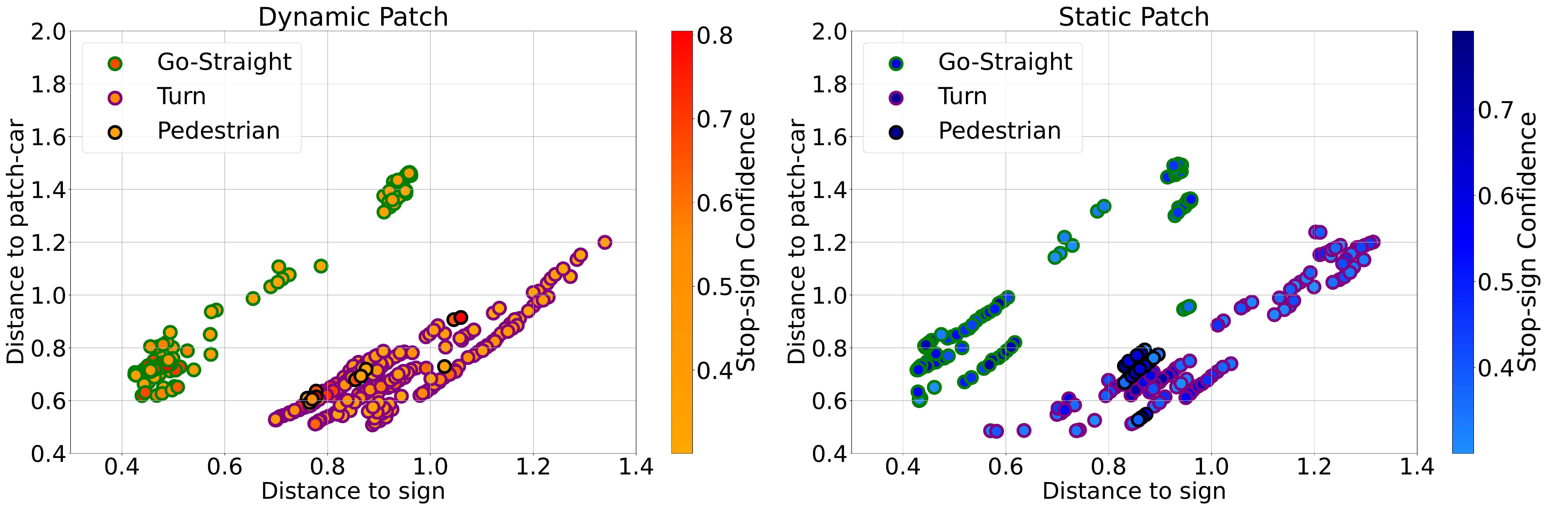}
    \caption{\rev{Scatter plots contrasting the effectiveness of dynamic and static patches in ``Stop-sign" adversarial attacks as a function of the distance (in meters) of \texttt{camera car} to both the sign and the \texttt{patch car} in the outdoor lighting condition.}}
    \label{fig:outdoor-hist}
\end{figure}

\rev{Additionally, to quantify the effectiveness more precisely, we generated scatter plots contrasting the performance of dynamic and static patches. Figure \ref{fig:outdoor-hist} provides a detailed view of the ``Stop-sign" adversarial attacks and their effectiveness as a function of the distance (in meters) between the camera car and both the sign and the patch car in outdoor lighting conditions.}

\begin{figure}[H]
    \centering
    \setlength{\lineskip}{4pt}
    \includegraphics[width=0.24\linewidth]{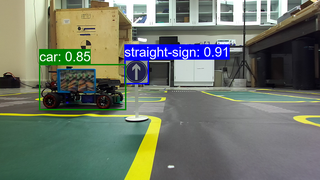}
    \includegraphics[width=0.24\linewidth]{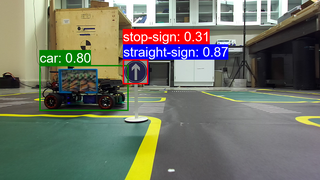}
    \includegraphics[width=0.24\linewidth]{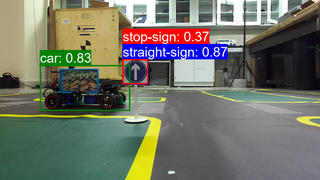}
    \includegraphics[width=0.24\linewidth]{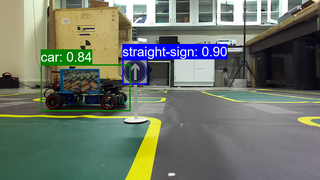}\\
    
    \includegraphics[width=0.24\linewidth]{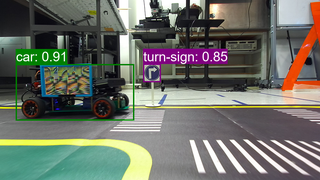}
    \includegraphics[width=0.24\linewidth]{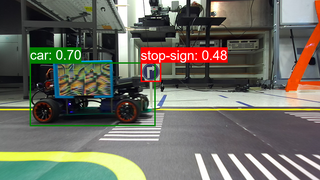}
    \includegraphics[width=0.24\linewidth]{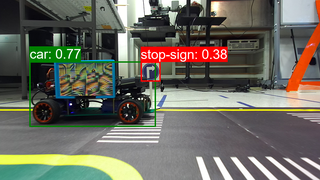}
    \includegraphics[width=0.24\linewidth]{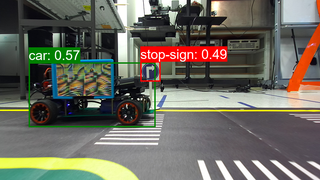}\\

    \includegraphics[width=0.24\linewidth]{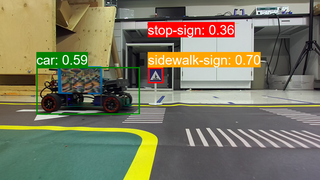}
    \includegraphics[width=0.24\linewidth]{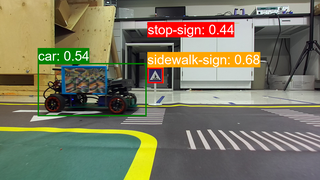}
    \includegraphics[width=0.24\linewidth]{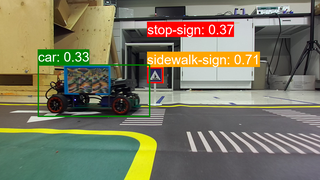}
    \includegraphics[width=0.24\linewidth]{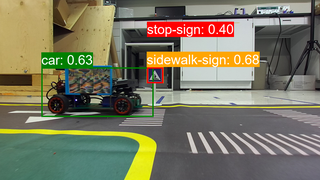}\\
    \caption{\rev{Sequential frames showing the performance of adversarial printed patch attacks. The frames depict that the confidence levels of \textbf{the misclassified ``Stop-sign" remain below the critical threshold of 0.5}, indicating the ineffectiveness of these patches in real-world conditions.}}
    \label{fig:print-test}
\end{figure}

\newpage
\rev{\subsection{Ineffectiveness of Printed Patches in Real-World Scenarios}
\label{sec:Appendix-printpatch}
We present the extended analysis of printed adversarial patches, highlighting the practical implications of the non-printability loss on real-world efficacy. The non-printability score, as discussed in \cite{Thys2019}, ensures that the colors utilized in the patches can be accurately reproduced by standard printers. However, this requirement often limits the overall effectiveness of the attack by restricting the range of colors that can be used.
 This constraint leads to a significant reduction in the effectiveness of the printed patches.

Figure~\ref{fig:print-test} illustrates that the confidence levels for the ``stop-sign" induced by these patches consistently fall below the critical threshold of 0.5. This outcome directly reflects the limitations imposed by the non-printability loss, which, while intended to make the patches printable, does not adequately mirror the appearance of real-world adversarial patches.}

\end{document}